\documentclass[a4paper,fleqn]{cas-sc}

\usepackage[authoryear]{natbib}
\usepackage{multirow}
\usepackage{makecell}
\usepackage{csvsimple}
\usepackage{graphicx}
\usepackage{array}
\usepackage{placeins}
\usepackage{xcolor}
\usepackage{soul}

\def\tsc#1{\csdef{#1}{\textsc{\lowercase{#1}}\xspace}}
\tsc{WGM}
\tsc{QE}

\begin{document}
\let\WriteBookmarks\relax
\def\floatpagepagefraction{1}
\def\textpagefraction{.001}

\shorttitle{}    

\shortauthors{}  

\title [mode = title]{Construction and Evaluation of Machine Learning Models for Near-Real-Time Fire Detection from MTG FCI Imagery}  

\tnotemark[1] 

\tnotetext[1]{} 

\author[1,2,3]{Asaf Vanunu}[orcid=0009-0007-6555-2817]
\ead{asafyu@post.bgu.ac.il}
\credit{Conceptualization, Methodology, Software, Validation, Formal analysis, Investigation, Data Curation, Writing - Original Draft, Writing - Review \& Editing, Visualization, Project administration}

\author[4]{Boaz Nadler}[orcid=0000-0002-9777-4576]
\ead{boaz.nadler@weizmann.ac.il}
\credit{Conceptualization, Methodology, Resources, Writing - Original Draft, Writing - Review \& Editing, Supervision, Project administration, Funding acquisition}

\author[2,3]{Arnon Karnieli}[orcid=0000-0001-8065-9793]
\cormark[1]
\ead{karnieli@bgu.ac.il}
\credit{Conceptualization, Methodology, Resources, Writing - Original Draft, Writing - Review \& Editing, Supervision, Project administration, Funding acquisition}

\affiliation[1]{organization={The Albert Katz International School for Desert Studies, The Jacob Blaustein Institutes for Desert Research, Ben-Gurion University of the Negev},
            addressline={Sde Boker Campus}, 
            city={Midreshet Ben-Gurion},
            postcode={8499000}, 
            country={Israel}}

\affiliation[2]{organization={The Goldman Sonnenfeldt School of Sustainability and Climate Change, Ben-Gurion University of the Negev},
            city={Beer Sheva},
            postcode={8410501}, 
            country={Israel}}

\affiliation[3]{organization={The Remote Sensing Laboratory, French Associates Institute for Agriculture and Biotechnology of Drylands, The Jacob Blaustein Institutes for Desert Research, Ben-Gurion University of the Negev},
            addressline={Sde Boker Campus}, 
            city={Midreshet Ben-Gurion},
            postcode={8499000}, 
            country={Israel}}

\affiliation[4]{organization={Department of Computer Science and Applied Mathematics, Weizmann Institute of Science},
            city={Rehovot},
            postcode={7610001}, 
            country={Israel}}

\cortext[1]{Corresponding author}


\begin{abstract}
Geostationary satellite observations are important for wildfire detection and monitoring. The current study evaluates machine learning models for MTG FCI near-real-time fire detection in 1- and 2-km spatial configurations and compares them with threshold-based algorithms. The models were trained and evaluated using VIIRS fire reference data across diverse ecological regions in Europe, Africa, and the Middle East. The key results are that 1-km models significantly outperform both their 2-km variants and operational threshold products. The constructed 1-km models achieved F1 scores higher by up to 0.36 compared to baseline products. Importantly, the 1-km models detected small fires with higher probability compared to competing models. Finally, the models robustly detected fires up to 260 min earlier than baseline products. To support open-source applications, our trained models are publicly available. 
\end{abstract}

\begin{graphicalabstract}
\includegraphics{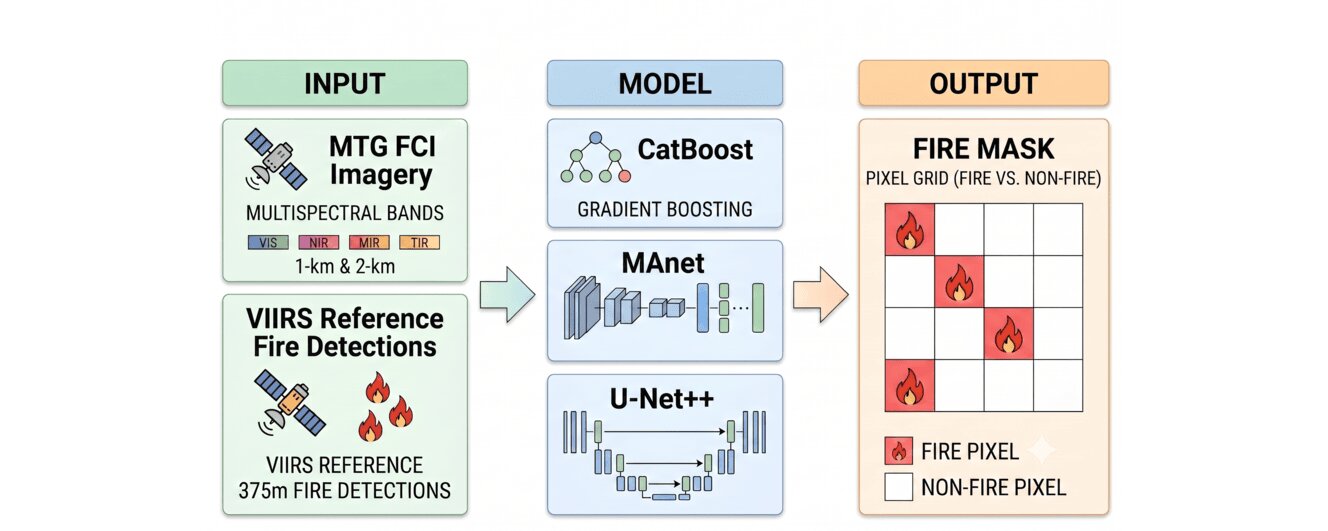}
\end{graphicalabstract}

\begin{highlights}
\item ML models were trained to detect fires using MTG FCI imagery.

\item 1-km models capture brightness temperature more accurately than 2-km variants.

\item 1-km ML models achieved F1 scores of 0.56 to 0.61, exceeding threshold products.

\item 1-km models detect small fires with higher probability than competing products.

\item On average, 1-km ML models detected large fires 32-70 min before threshold products.
\end{highlights}


\maketitle


\section{Introduction}\label{sec:intro}

Wildfires are a natural hazard that poses a severe threat to human life, ecosystems, and economies. Over the past three decades, the frequency of extreme fire events has increased dramatically \citep{cunningham2025climate, Vazquez_2025_fire_weather_synchronicity}. Recent extreme wildfire events occurred during the 2025 and 2026 European wildfire season. Severe weather conditions triggered widespread fire outbreaks that ultimately consumed massive areas and caused serious socio-economic impacts \citep{sanchez_europe_2025_fires}. Given the devastating scale of these intense fires, timely and accurate fire detection is crucial for rapid response and prevention of small fires from escalating into uncontrollable disasters.

One common method for wildfire detection and monitoring is based on satellite imagery \citep{Ghali_2023_fire_detection_review}. Specifically, two types of space systems are used for fire detection and monitoring: Low Earth Orbit (LEO) and Geostationary (GEO), both have designated algorithms. Depending on the specific satellite and sensor configuration, LEO satellites offer high spatial resolution, ranging from 30 to 1000 m, with revisit times varying from 12 hours to several days \citep{Giglio_MODIS_2003, WOOSTER_Sentinel_3_SLSTR_2012, SCHROEDER_2014_VIIRS_active_fire_detection, SCHROEDER_Landsat_2016, Hu_S2_fire_detection_2021}. However, their low temporal resolution restricts their utility for early fire detection. Conversely, GEO satellites rotate at the same angular velocity as the Earth and continuously monitor the same region at a continental scale. This property enables high temporal resolution, with image acquisition occurring every 5 to 15 min \citep{Schmit_GOES_closer_look_2017, Da_2015_Himawari_imager}. However, this comes at the cost of spatial resolution, which is coarser than the resolution of LEO platforms. In particular, for most GEO satellites the spectral bands relevant for fire detection are acquired at 2-km spatial resolution; hence, detecting small fires before they expand is a major challenge.

In this context, the next-generation GEO satellite, i.e., the Meteosat Third Generation Imager-1 (MTG-I1)  \citep{Holmlund_2021_MTG_mission}, holds great potential to improve early fire detection. Launched in December 2022 and fully operational since early 2024, MTG covers Europe, Africa, and the Middle East. 
Equipped with the Flexible Combined Imager (FCI)  \citep{Ouaknine_2017_FCI_on_board_MTG}, it operates in both 1-km and 2-km spatial configurations and provides full disk images every 10 minutes. Specifically, in the spectral bands critical for fire detection ($3.8\ \mu\text{m}$ and $10.5\ \mu\text{m}$), the HR configuration has a spatial resolution of 1-km, while the NR configuration provides 2-km resolution. This configuration improves on the legacy Spinning Enhanced Visible and Infrared Imager (SEVIRI) onboard the Meteosat Second Generation (MSG) satellite \citep{Sifakis_2011_MSG_fire_detection}. In comparison, SEVIRI has a 3-km spatial resolution and a 15 min scan time.

The MTG FCI has two fire detection products: the 2-km Active Fire Monitoring (FIR) product \citep{FIR_ATBD_2019} and the 1-km Fire Radiative Power (FRP-PIXEL) product \citep{Wooster_2015_FRP}. The FRP-PIXEL product, originally developed for legacy SEVIRI imagery, has been adapted to MTG FCI. This product was used to compare the fire detection capabilities of MTG FCI and lower-resolution MSG SEVIRI \citep{XU_2026_MTG_FCI_fire_improve}. Their findings were that with the higher resolution MTG FCI images, the fire detection capabilities were much improved compared to MSG SEVIRI, and fire could be detected up to 4 hours earlier.

Both the FRP-PIXEL and FIR products are based on threshold and contextual algorithms that follow predefined rules. While effective, these strict decision rules are over-conservative, leading to high omission rates \citep{Wickramasinghe_Himawari_FSA_2020, XU_FRP_GOES_2021}.
In particular, \citet{XU_2026_MTG_FCI_fire_improve} reported omission rates up to 68\% for the MTG FRP-PIXEL product when compared to fire detections from LEO instruments. 

An approach to improve the accuracy of GEO fire detections is to apply machine learning (ML) algorithms \citep{YANG_2024_review_fire_ML_RS}. These models learn complex, non-linear patterns from multispectral satellite imagery. ML models require annotated datasets for training. Typically, reference data derive from national forestry databases \citep[e.g.,][]{Jang_2019_ML_H8_korea, Kang_2022_DL_H8_latency} or higher-resolution LEO active fire products \citep[e.g.,][]{Zhao_2022_DL_GOES, Zhang_2023_RF_H8_VIIRS}. The latter necessitates temporally matched GEO and LEO pairs covering the same region. By training on large annotated datasets, ML models can generalize across diverse environmental conditions. For example, \citet{KANG_2023_Toward_H8} trained a Convolutional Neural Network (CNN) on images from the Advanced Himawari Imager (AHI) onboard the Japanese Himawari GEO satellite. They reported lower omission errors and overall higher detection accuracy than the operational threshold-based Himawari fire product. In addition, their model was applied to regions outside the training area, demonstrating strong generalization capabilities. Recently, \citet{Mercatini_MTG_FCI_fire_detection_2026} applied a deep learning (DL) anomaly detection approach to MTG FCI imagery. Their approach achieved higher overall accuracy and improved detection capabilities compared to threshold-based methods.

Building on the above GEO fire detection studies, our overarching goal is to train ML models for wildfire detection using MTG FCI imagery. We aim to outperform current threshold-based products and make our models publicly available. This goal, alongside the improved spatial resolution of the MTG FCI, raises the following questions:
\begin{enumerate}
\item \textbf{Advantage of HR over NR:} To what extent does the 1-km HR imagery of the MTG FCI enable improved fire detection capabilities compared to its 2-km NR?
\item \textbf{Advantage of ML over threshold-based:} To what extent do the trained ML algorithms improve upon traditional threshold-based frameworks on a multi-continental scale?
\item \textbf{Detection of small fires:} How do these advanced algorithms improve the capability to detect small fires?
\item \textbf{Early detection of fire onset:} To what extent do the trained ML models identify fire events earlier than threshold-based algorithms?
\end{enumerate}
To study these questions, this study trains and evaluates several ML models for MTG FCI fire detection on a multi-continental scale. Specifically, we quantify the gains of ML models trained and applied to the 1-km HR configuration over the 2-km NR. Furthermore, we compare these models to MTG FCI threshold-based products in terms of overall accuracy, detection probability, and detection latency. 
To this end, we constructed a large annotated dataset of MTG FCI images containing over 150,000 fire pixels. We used this dataset to train several ML models at 1-km and 2-km spatial resolutions. The evaluated algorithms included gradient boosting, Multi-Attention Network (MAnet), and Nested U-Net (U-Net++). The reference fire detections for model training and evaluation were derived from the LEO Visible Infrared Imaging Radiometer Suite (VIIRS) 375 m fire product. The training area spans vast, ecologically diverse regions in Western Europe, Eastern Africa, and the Eastern Mediterranean. To assess generalization, the models were evaluated on an independent dataset covering parts of Eastern Europe, the Middle East, and West Africa.

\section{Study Area and Data sources}\label{sec:data and study area}
This section outlines the study regions selected for model training and evaluation (Section \ref{subsec:study_area}), and the datasets used in this study (Section \ref{subsec:data_sources}). Specifically, the LEO fire detections used as reference data for labeling and validation are detailed in Section \ref{subsubsec:VIIRS_reference_data}. In addition, the MTG FCI multispectral imagery and its benchmark threshold-based products are described in Section \ref{subsubsec:MTG_FCI_products}.
Finally, Section \ref{subsubsec:fire_events_databse} outlines the fire event database compiled for the early detection analysis.

\subsection{Study Area}\label{subsec:study_area}
\begin{figure}
    \centering
    \includegraphics[width=.8\textwidth]{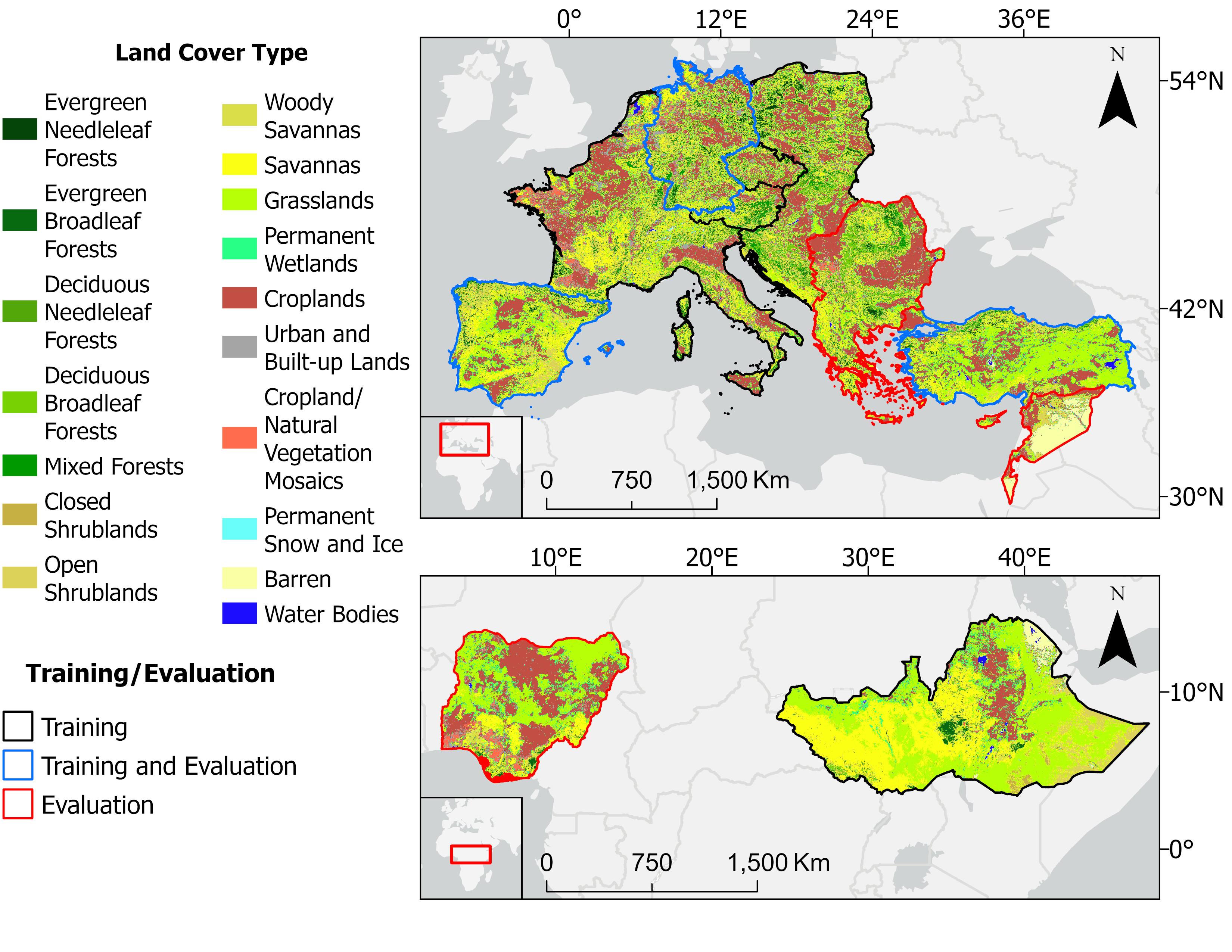}
    \caption{Study areas over land cover maps, covering Europe and Central Africa. The boundaries of training regions are delinted by black line, the evaluation regions by red, and the regions used for both are marked in light blue.}
    \label{fig:study area}
\end{figure}

The study area spans a vast geographic domain across diverse ecosystems (Figure \ref{fig:study area}). It encompasses forests, croplands, shrublands, savannas, urban areas, and barren deserts with varied climatic conditions. This environmental diversity was selected to ensure the model's robustness and assess its generalization capabilities.

The study area is partitioned into distinct regions, each assigned to one of three categories:

\begin{enumerate}
    \item \textbf{Regions used exclusively for training:} These comprise parts of Western and Central Europe, as well as Ethiopia and South Sudan in Africa.
    \item \textbf{Regions used exclusively for evaluation:} These include the Balkans, the Middle East (specifically Cyprus, Syria, and Israel), and Nigeria.
    \item \textbf{Regions used for both training and evaluation:} These include Spain, Portugal, Germany, and Turkey.
\end{enumerate}

In total, the training regions cover approximately 6.1 million~km\textsuperscript{2}, while the independent evaluation regions cover 3.4 million~km\textsuperscript{2}.

\subsection{Data sources}\label{subsec:data_sources}
To train and evaluate the ML models, data from two satellite platforms were downloaded:
\begin{enumerate}
\item LEO VIIRS 375 m fire detections, which serve as the reference for MTG FCI fire pixel annotation and model evaluation.
\item MTG FCI multispectral imagery and its products, which serve as model inputs and evaluation benchmark.
\end{enumerate}

MTG FCI products were acquired within a $\pm 3$ min window from corresponding VIIRS fire detections. The dataset period was dictated by MTG FCI product availability. The training dataset spans February to December 2025. The evaluation dataset covers July 2025 to March 2026, following the availability of the FIR product. 

To prevent data leakage, the training and evaluation datasets are independent. Some months were used for both training and evaluation. For these months, different calendar days were assigned to each dataset. In total, the training dataset comprises 457,747 VIIRS fire detections corresponding to 1,798 MTG FCI images. The evaluation dataset contains 100,876 VIIRS fire pixels and is summarized in Table \ref{tab:dataset_summary}.

\begin{table}
\centering
\caption{Summary of the evaluation dataset.}
\label{tab:dataset_summary}
\begin{tabular}{llcc}
\hline
\textbf{Region} & \textbf{Months} & \textbf{\makecell[c]{Number of\\ MTG FCI Images}} & \textbf{\makecell[c]{Number of VIIRS\\ Fire Detections}} \\ \hline
Balkan & \multirow{5}{*}{July, August, September 2025} & 159 & 8,387 \\
Germany & & 138 & 11,125 \\
Middle East & & 109 & 3,190 \\
Spain - Portugal & & 143 & 15,197 \\
Turkey & & 151 & 20,702 \\ \hline
Nigeria & January, February, March 2026 & 155 & 42,275 \\ \hline
\end{tabular}
\end{table}

\subsubsection{VIIRS Fire Product}\label{subsubsec:VIIRS_reference_data}
To annotate MTG FCI fire pixels and evaluate model performance, outputs of the VIIRS fire products were downloaded. Specifically, 375-m VNP14IMG and VJ114IMG Level 2 products were obtained from the Suomi-NPP and NOAA-20 satellites \citep{SCHROEDER_2014_VIIRS_active_fire_detection}.
These threshold-based products use the 3.7 $\mu\text{m}$ and 11.4 $\mu\text{m}$ bands as input to detect thermal anomalies. The output is a pixel-level mask flagging detected fire pixel locations. Fire pixels flagged as low-confidence detections were omitted from the dataset. The data were downloaded from NASA LAADS DAAC\footnote{NASA LAADS DAAC: \url{https://ladsweb.modaps.eosdis.nasa.gov/}, accessed on April 1, 2026.} and NASA Earthdata.\footnote{NASA Earthdata Search: \url{https://search.earthdata.nasa.gov/}, accessed on April 1, 2026.}

\subsubsection{MTG FCI Products}\label{subsubsec:MTG_FCI_products}
Four bands from MTG FCI Level 1C multispectral imagery acquired in both 1-km and 2-km configurations served as input for the ML models. Specifically, the selected bands (detailed in Table \ref{tab:fci_bands}) include the Visible 0.6 $\mu$m (VIS), Near Infrared 2.2 $\mu$m (NIR), 3.8 $\mu$m Mid-Wave Infrared (MIR), and Thermal Infrared 10.5 $\mu$m (TIR). These bands were selected since they are available in both HR and NR configurations.
\begin{table}
\centering
\caption{Selected MTG FCI spectral bands used as ML models input.}
\label{tab:fci_bands}
\begin{tabular}{lcc}
\hline
\textbf{Band ($\mu$m)} & \textbf{HR (km)} & \textbf{NR (km)} \\ \hline
VIS 0.6 & 0.5 & 1.0 \\
NIR 2.2 & 0.5 & 1.0 \\
MIR 3.8 & 1.0 & 2.0 \\
TIR 10.5 & 1.0 & 2.0 \\ \hline
\end{tabular}
\end{table}

To exclude cloud-covered pixels, cloud masking was applied using the 2-km Level 2 Cloud Mask (CLM) product \citep{mtg_fci_clm_atbd_2015}. It receives multispectral radiance and reflectance values as inputs. The output is a mask flagging whether a pixel is cloudy or clear.
To benchmark the ML models, two threshold-based fire products were used: the European Organisation for the Exploitation of Meteorological Satellites (EUMETSAT) Level 2 FIR product at 2-km resolution \citep{FIR_ATBD_2019}; and the Land Surface Analysis Satellite Application Facility (LSA SAF) FRP-PIXEL product at 1-km resolution \citep{Wooster_2015_FRP}. The input for both products are the 3.8 $\mu$m and 10.5 $\mu$m brightness temperatures (BT) values. The output of these products is a pixel-level mask indicating the presence of fire.

The MTG FCI multispectral imagery, CLM, and FIR products were downloaded from the EUMETSAT Data Store\footnote{EUMETSAT Data Store: \url{https://data.eumetsat.int/}, accessed on April 1, 2026.}. The FRP-PIXEL product was acquired from the LSA SAF website\footnote{EUMETSAT LSA SAF: \url{https://lsa-saf.eumetsat.int/}, accessed on April 1, 2026.}.

\subsubsection{Fire Events Database}\label{subsubsec:fire_events_databse}
\begin{figure}
\centering
\includegraphics[width=.7\textwidth]{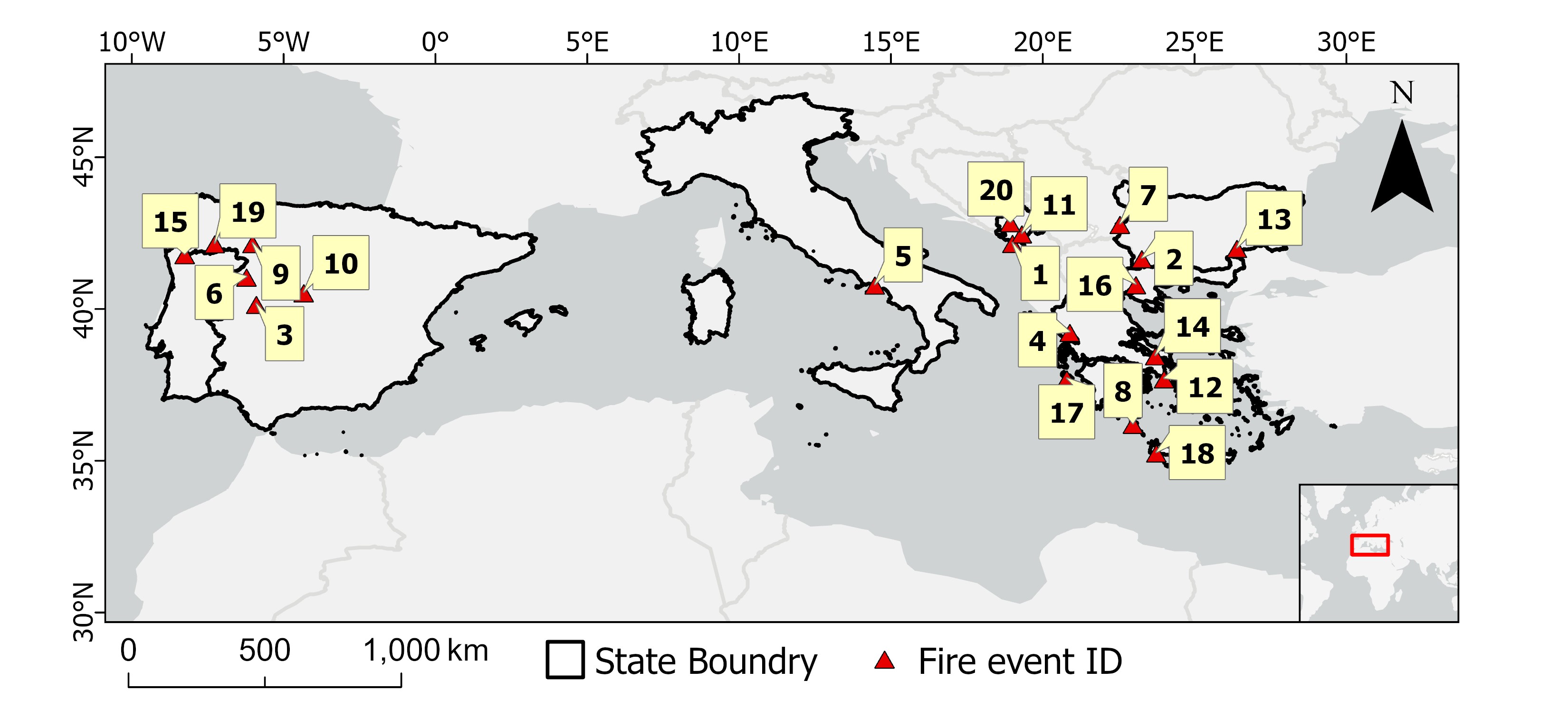}
\caption{Locations of 20 fire events across Europe during summer 2025.}
\label{fig:fire_event_map}
\end{figure}

Fire perimeter data were downloaded to evaluate early detection performance. These data covered 20 fire events from the summer of 2025 in Italy, Spain, Portugal, Bulgaria, Montenegro, and Greece (Figure \ref{fig:fire_event_map}). Out of the 20 fire events, 14 occurred outside the training region. The remaining six events occurred on dates that were not used for training. Fire perimeter data were obtained from the Copernicus Emergency Management Service (EMS).\footnote{Copernicus Emergency Management Service: \url{https://mapping.emergency.copernicus.eu/}, accessed on April 1, 2026.}
Additional information, such as fire locations, EMS ID, and dates, is detailed in Appendix \ref{app:fire_events_list}. 

\section{Methodology}\label{sec:methodology}
This section outlines the workflow for training (Section \ref{subsec:model_training}) and evaluating (Section \ref{subsec:evaluation process}) ML fire-detection models for MTG FCI imagery. First, spatial alignment between MTG FCI imagery and VIIRS reference fire detections is outlined. Next, we detail the construction of the annotated fire-non-fire dataset for model training. Finally, the evaluation framework for the models is presented.     

\subsection{Data Alignment Between MTG FCI and VIIRS}\label{subsec:data_alignment}
To construct the annotated training dataset, all MTG FCI spectral bands are mapped to a uniform spatial grid. Furthermore, because MTG FCI and VIIRS have different coordinate systems and spatial resolutions, spatial alignment between the two instruments is critical. First, MTG FCI bands were resampled to a uniform grid with a target MIR 3.8 $\mu$m band resolution: 1-km for the HR and 2-km for the NR configuration. Resampling was performed using the Satpy Python library \citep{Satpy_2026}. Next, the VIIRS fire detections were remapped onto the two corresponding MTG FCI HR and NR grids. Within these grids, pixel values represent the total number of intersecting VIIRS fire centroids. Pixels with no intersecting VIIRS fire detections were assigned a value of 0.

\subsection{Labeled Dataset Construction}\label{subsec:dataset_construction}
Following the remapping of the VIIRS fire detections onto the MTG FCI grids, two distinct labeled datasets were constructed: 
\begin{enumerate}
    \item tabular dataset for gradient boosting model training, detailed in Section \ref{subsubsec: ML dataset},
    \item Patch-based dataset for DL model training, detailed in Section \ref{subsubsec DL dataset}.
\end{enumerate}
   For all ML frameworks, separated datasets were constructed for the 1-km and 2-km MTG FCI imagery configurations.

In the construction of these annotated datasets, a key issue is to account for spatial misalignments between LEO and GEO fire detection locations. This spatial misalignment is caused by differences in satellite viewing angles, parallax effect, and other discrepancies \citep{PESTANA_parallax_2022, Ayala_Parallax_2023, Vanunu2025_FireMisalignments}. These factors can introduce a spatial shift of several kilometers between GEO and LEO fire detections. This can lead to noisy pixel labeling that can reduce classifier accuracy \citep{Natarajan_noisy_labels_2013}. To mitigate these misalignments, a spatial buffer is commonly applied around the GEO/LEO fire detection; this approach is widely used for evaluation of GEO fire products \citep{Freeborn_validation_2014, Wickramasinghe_Himawari_FSA_2020, HALL_2019_GOES_SEVIRI, Hall_2023_GOES_H8, XU_2026_MTG_FCI_fire_improve, Liu_Fusing_context_H8_2026}. 
Accordingly, a $3 \times 3$ pixel buffer window was applied to label the MTG FCI fire pixels. 
An MTG FCI Fire Index (FI) was calculated using the normalized ratio between the MIR band at 3.8~µm and the TIR band at 10.5~µm:
\begin{equation}
FI = \frac{B_{3.8} - B_{10.5}}{B_{3.8} + B_{10.5}}
\label{eq:fire_index}
\end{equation}
where $B_{3.8}$ and $B_{10.5}$ represent the BT in Kelvin (K). Since the 3.8~µm band is sensitive to fires while the 10.5~µm band is not, high FI values serve as a reliable indicator of active fire presence \citep{Zhao_2022_DL_GOES}.

Following the FI computation, the labeling pipeline was executed as follow: let $V_{rm}$ be the remapped VIIRS, where a cell value of $V_{rm}(i,j) \geq 1$ indicates the presence of at least one VIIRS fire detection centroid within that specific MTG FCI grid cell $(i,j)$. Let $FCI(i,j)$ represent the corresponding pixel in the MTG FCI image at the same timestamp. For every $V_{rm}(i,j) \geq 1$, a $3 \times 3$ buffer was applied around $FCI(i,j)$. Within this buffer window, the MTG FCI pixel with the highest FI value was labeled as a fire pixel. 

The strategy for labeling non-fire instances differed between the tabular and patch-based datasets. For the tabular dataset, non-fire labels were assigned to 500 randomly sampled pixel locations per image, provided they were not adjacent to any VIIRS fire detections. For the patch-based dataset, non-fire patches were selected from areas without VIIRS fire detections.

\subsubsection{Tabular Dataset}\label{subsubsec: ML dataset}
For each labeled pixel (fire or non-fire) in the dataset, the input features comprise its pixel-level values alongside surrounding contextual data. The pixel-level features include the band values from the four selected MTG FCI spectral bands and the computed FI. The contextual features consist of five statistical metrics (mean, median, minimum, maximum, and standard deviation) calculated within a $3 \times 3$ spatial window, excluding the center pixel itself. These five metrics are computed independently for each of the five bands (the four spectral bands and the FI), yielding 25 contextual features per pixel. Pixels flagged as cloud-covered by the CLM product were excluded from these calculations. If an entire $3 \times 3$ neighborhood was covered by clouds, the contextual features were assigned a fill value of -999. In total, there are 30 input features per observation.

To construct accurate ML fire detection models, training data filtering criteria were applied. Only labeled fire pixels with $FI>0$ were retained. Furthermore, fire pixels were retained only if their corresponding $V_{rm}$ remapped grid cell contained at least two intersecting VIIRS fire detections. This threshold was selected because models trained on at least two detections yielded the best performance. Details regarding the resulting training datasets appears in Table \ref{tab:ml_dataset_summary}.

\begin{table}
\centering
\caption{Summary of the tabular training datasets.}
\label{tab:ml_dataset_summary}
\begin{tabular}{lcc}
\hline
\textbf{Resolution} & \textbf{Fire Pixels} & \textbf{Non-Fire Pixels} \\ \hline
1-km (HR) & 82,466 & \multirow{2}{*}{981,000} \\
2-km (NR) & 75,186 & \\ \hline
\end{tabular}
\end{table}

\subsubsection{Patch-based Dataset}\label{subsubsec DL dataset}
The DL training dataset consists of tensors with dimensions: $128 \times 128 \times 4$, where the last dimension corresponds to the four selected MTG FCI spectral bands. Each input patch is paired with a corresponding $128 \times 128 \times 1$ binary reference mask. Within this mask, a pixel value of 1 denotes an active fire, while 0 denotes background. A summary of the patches and individual pixel labels for the 1-km and 2-km datasets is provided in Table \ref{tab:dl_dataset_summary}.

\begin{table}
\centering
\caption{Summary of the patch-based datasets.}
\label{tab:dl_dataset_summary}
\begin{tabular}{lcccc}
\hline
\textbf{Resolution} & \textbf{Fire Patches} & \textbf{Non-Fire Patches} & \textbf{Fire Pixels} & \textbf{Non-Fire Pixels (Million)} \\ \hline
1-km (HR) & 14,174 & 5,515 & 153,620 & 325 \\
2-km (NR) & 7,221 & 1,850 & 113,847 & 129 \\ \hline
\end{tabular}
\end{table}

\subsection{Models Training}\label{subsec:model_training}
In this study, three distinct predictive models were trained and evaluated:
First, a gradient boosting algorithm, CatBoost (detailed in Section~\ref{subsubsec: Catboost training}). Next, two hybrid DL architectures, MAnet and U-Net++ (described in Section~\ref{subsubsec: DL training}). The trained models, source code, and implementation examples are publicly available in a GitHub repository.\footnote{\url{https://github.com/asafvanunu/Github_repo_MTG_ML_DL_fire_detection}}

All model training and evaluation pipelines were executed on a 64-bit workstation equipped with an Intel(R) Core(TM) i7-10700 CPU running at 2.90 GHz, 64.0 GB of RAM, and an NVIDIA GeForce RTX 2080 SUPER GPU with 8 GB of VRAM. The computational environment was built on Python version 3.9. The CatBoost model was trained using CatBoost library version 1.2.8. The MAnet and U-Net++ models were trained using the Segmentation Models PyTorch library version 0.5.0 \citep{Iakubovski_SMP_2019}.

\subsubsection{CatBoost}\label{subsubsec: Catboost training}
CatBoost (Categorical Boosting) is a gradient boosting framework that constructs an ensemble of decision trees \citep{catboost}. It is widely applied in various domains \citep{hancock2020catboost}, including remote sensing \citep{boosting_in_RS}. 
CatBoost introduces advantages over traditional gradient boosting methods. Specifically, it reduces data leakage during model training and uses symmetric decision trees to prevent overfitting. For the training setup, the dataset was randomly partitioned into 80\% training and a 20\% validation set. Details regarding the hyperparameter configurations are provided in Appendix~\ref{app:CatBoost Hyperparameters}.

\subsubsection{Deep Learning Architectures}\label{subsubsec: DL training}
Two hybrid DL architectures were trained in this study. These architectures allow custom pairing of different encoders and decoders. 
The first configuration pairs a Multi-Attention Network (MAnet) decoder \citep{Fan_MA-Net_2020} with a Mix Transformer (MiT) encoder backbone (CNN-Transformer hybrid). The second configuration combines a U-Net++ decoder \citep{Zhou_Unet++_2018} with a Squeeze-and-Excitation Residual Network (SE-ResNet) encoder \citep{Hu_2018_SEresnet} (CNN hybrid).
Using hybrid models overcomes the limitations of standard homogeneous networks \citep{Li_Segmentation_RS_2024}.
Standard CNNs struggle to capture long-range contextual dependencies. Conversely, Transformers capture global relationships across the entire image but can miss fine-grained local features. Hybrid architectures combine the strengths of both \citep{chen_transunet_2021, WANG_FireVitnet_2024}.

For the two models, extreme class imbalance between fire and non-fire pixels was addressed using a Focal Loss function. Applied during training, the function down-weights non-fire background pixels. Consequently, the optimization process focuses on learning scarce fire pixels. Further details regarding model initialization, learning rate, and stopping criteria are provided in Appendix \ref{app:DL Hyperparameters}.

\subsection{Model Accuracy Evaluation}\label{subsec:evaluation process}
To evaluate the accuracy of a fire detection algorithm, spatial misalignments between MTG FCI and VIIRS fire detections were considered. 
Similar to Section \ref{subsec:dataset_construction}, let $FCI_m(i,j)=1$ denote that at FCI pixel location $(i,j)$ the fire detection algorithm predicted a fire.  
A pixel $(i,j)$ with $FCI_m(i,j)=1$ is classified as a True Positive (TP) if there is at least one $V_{rm}(i',j')\geq1$ within a $3 \times 3$ window around $(i,j)$.
Otherwise, the pixel is considered a False Positive (FP). 
A False Negative (FN) pixel is defined as follows.
If the $3 \times 3$ window around $V_{rm}(i,j)\geq1$ does not contain a model fire detection, then the pixel is defined as FN.

The accuracy of a fire detection algorithm is evaluated by its recall, precision, and F1 scores.
These are standard metrics for classifier assessment.
In the context of fire detection, recall is the proportion of fire pixels that the algorithm correctly identifies as fires.
Precision is the ratio of correctly detected fire pixels to the total number of flagged fire pixels by the algorithm.
The harmonic mean of precision and recall is the F1 score. All three metrics range from 0 to 1, where higher values indicate better detection accuracy.

To compute recall, precision, and F1, the total number of TP, FP, and FN pixels was summed across the entire image dataset.
We denote these quantities by $\text{TP}^\text{total}$, $\text{FP}^\text{total}$ and $\text{FN}^\text{total}$. Accordingly, let $n$ denote a pair of MTG FCI/remapped VIIRS images, and let $i$ represent the index of a single image. For each image $i$, we denote by $TP_i$, $FP_i$, and $FN_i$ the number of pixels in that image classified as TP, FP, and FN, respectively.
The totals were then calculated:   
\begin{equation*}
    \text{TP}^{\text{total}} = \sum_{i=1}^{n} \text{TP}_i \qquad \text{FP}^{\text{total}} = \sum_{i=1}^{n} \text{FP}_i \qquad \text{FN}^{\text{total}} = \sum_{i=1}^{n} \text{FN}_i
\end{equation*}
Based on these totals, the precision, recall, and F1 metrics were calculated:
\begin{align}
    \text{Recall} &= \frac{\text{TP}^{\text{total}}}{\text{TP}^{\text{total}} + \text{FN}^{\text{total}}} \\
    \text{Precision} &= \frac{\text{TP}^{\text{total}}}{\text{TP}^{\text{total}} + \text{FP}^{\text{total}}} \\
    \text{F1} &= \frac{2 \times \text{Precision} \times \text{Recall}}{\text{Precision} + \text{Recall}}
\end{align}

\section{Results}\label{sec:results}
This section evaluates the 1-km and 2-km MAnet, U-Net++, and CatBoost models, and compares their performance with the FIR and FRP-PIXEL products. First, to assess sensitivity to fire, Section~\ref{subsec: Models Distributions} details the distribution of TP $\mathit{BT}_{3.8\,\mu\text{m}}$ values. Section~\ref{subsec:overall_performance} then assesses model performance using recall, precision, and F1 scores. Following this, Section~\ref{subsec:probability_of_detection} examines the probability of fire detection as a function of fire size. Finally, Section~\ref{subsec:early_detection} analyzes the early detection capabilities of the models across 20 fire events.

\subsection{Brightness Temperature Distributions of Fire-Detected Pixels}
\label{subsec: Models Distributions}
\begin{figure}
    \centering
    \includegraphics[width=.7\textwidth]{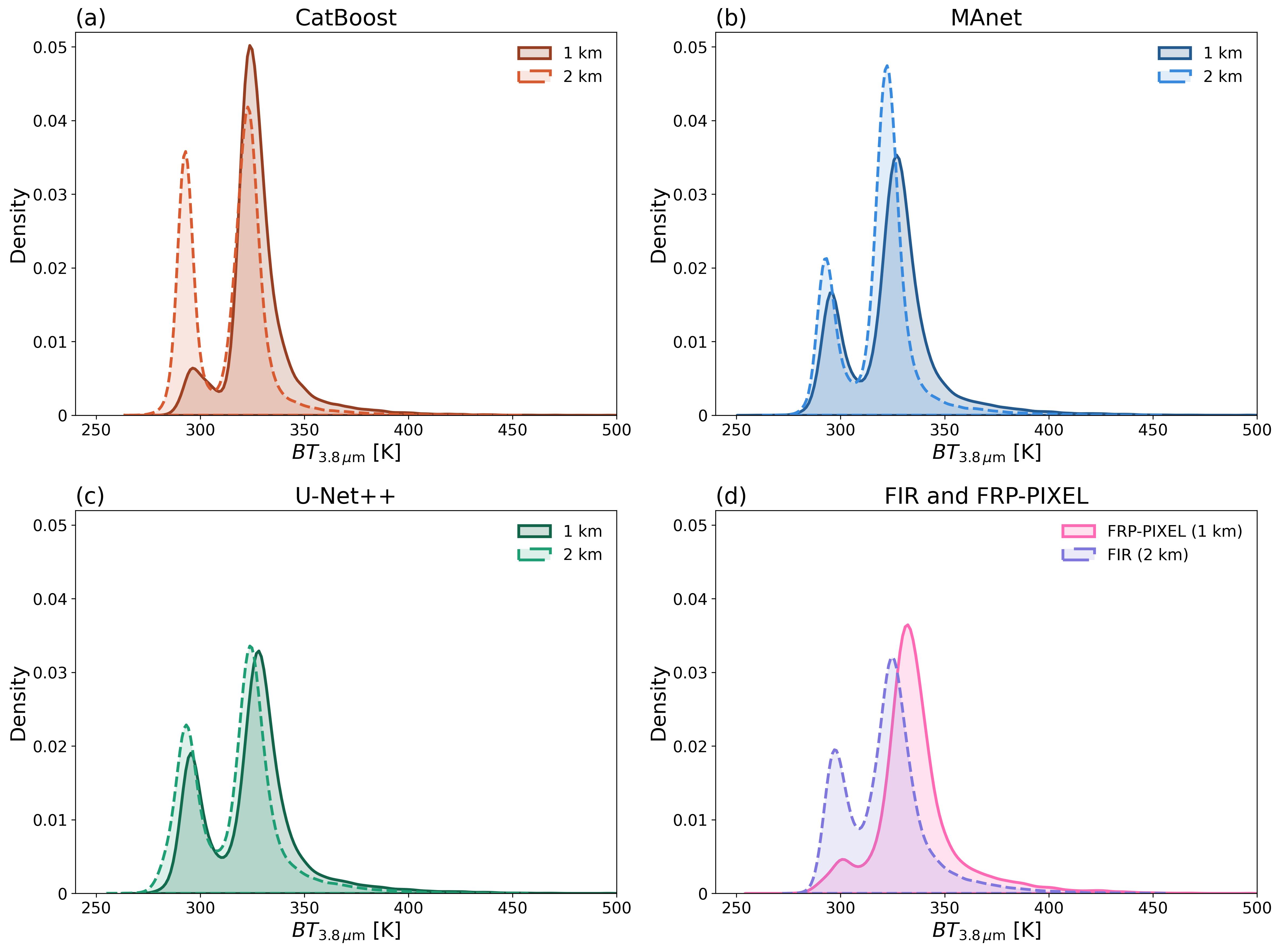}
    \caption{Probability density distributions of $\mathit{BT}_{3.8\,\mu\text{m}}$ for TP fire detections compared between 1 and 2-km model variants of (a) CatBoost, (b) MAnet, (c) U-Net++, and (d) the FRP-PIXEL and FIR products. For each model, a two-sample Kolmogorov-Smirnov test confirmed a statistically significant difference between the 1-km and 2-km distributions ($p < 0.001$).}
    \label{fig:model_grid_distribution}
\end{figure}

To evaluate sensitivity to fire, probability density distributions of $\mathit{BT}_{3.8\,\mu\text{m}}$ values for TP fire-detected pixels were compared between the 1-km and 2-km model variants (Figure~\ref{fig:model_grid_distribution}a–c). Additionally, the 1-km FRP-PIXEL product was compared to the 2-km FIR product (Figure~\ref{fig:model_grid_distribution}d).

All models have bimodal probability density distributions. The right peak corresponds to high-temperature fire pixels centered around $\sim325$~K. The left peak near $\sim295$~K represents lower-temperature fire detections.
The 1-km model distributions shift toward warmer temperatures, whereas the 2-km distributions shift toward cooler temperatures.
This shift occurs because the 2-km resolution covers a larger surface area, incorporating more non-fire background compared to the 1-km grid. Consequently, the 1-km resolution captures fire BT values more accurately.
A two-sample Kolmogorov-Smirnov test confirmed that the $\mathit{BT}_{3.8\,\mu\text{m}}$ distributions of fire detections differed significantly between the 1 and 2-km models ($p < 0.001$).

\begin{figure}
    \centering
    \includegraphics[width=.9\textwidth]{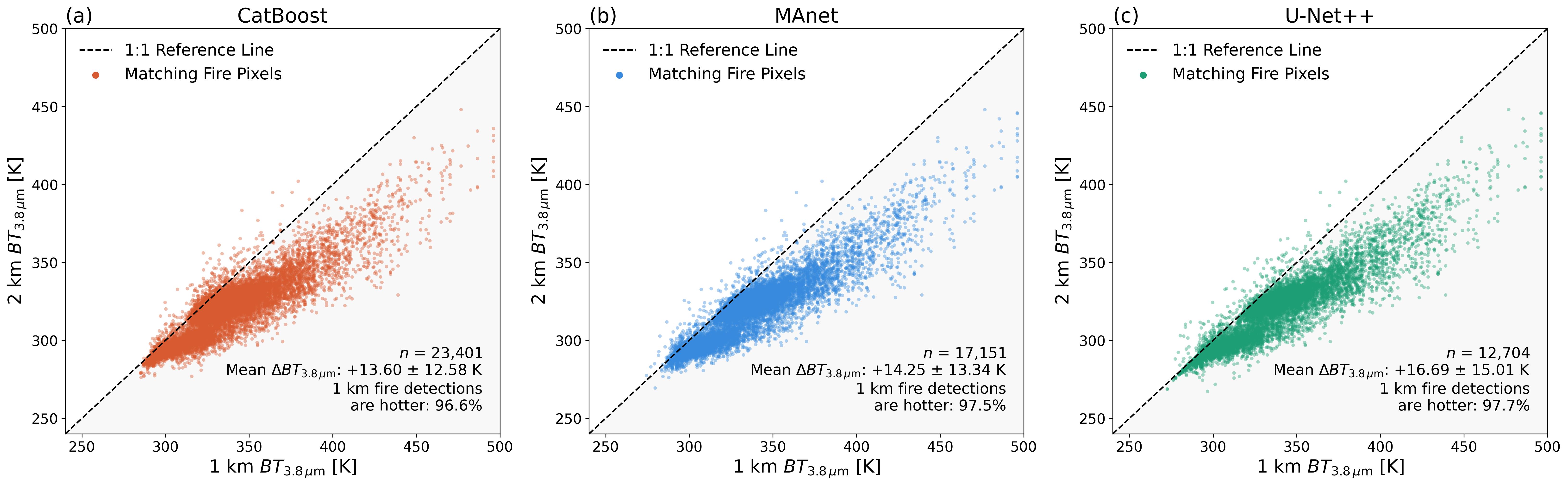}
    \caption{Scatter plots comparing pixel-matched $\mathit{BT}_{3.8\,\mu\text{m}}$ for TP fire detections between 1-km and 2-km variants of (a) CatBoost, (b) MAnet, and (c) U-Net++.}
    \label{fig:paired_metrics_comparison}
\end{figure}

In a separate analysis, direct pixel-matched TP fire detection $\mathit{BT}_{3.8\,\mu\text{m}}$ values were compared between resolutions (Figure~\ref{fig:paired_metrics_comparison}).
This comparison further confirmed that the 1-km resolution captures BT values more accurately and yields higher detected temperature. For CatBoost, MAnet, and U-Net++, over 96.6\% of pixel-matched detections showed higher BT at 1-km resolution, with mean $\Delta\mathit{BT}_{3.8\,\mu\text{m}}$ differences exceeding +13~K. 

\subsection{Overall Model Performance Comparison}
\label{subsec:overall_performance}

Overall model performance in all evaluation regions was assessed using recall, precision, and F1 scores. Metrics were evaluated for the 1 and 2-km models alongside the FIR and FRP-PIXEL threshold-based products. An additional region-specific evaluation is provided in Appendix~\ref{app:regional_performance}.

\begin{figure}
    \centering
    \includegraphics[width=.7\textwidth]{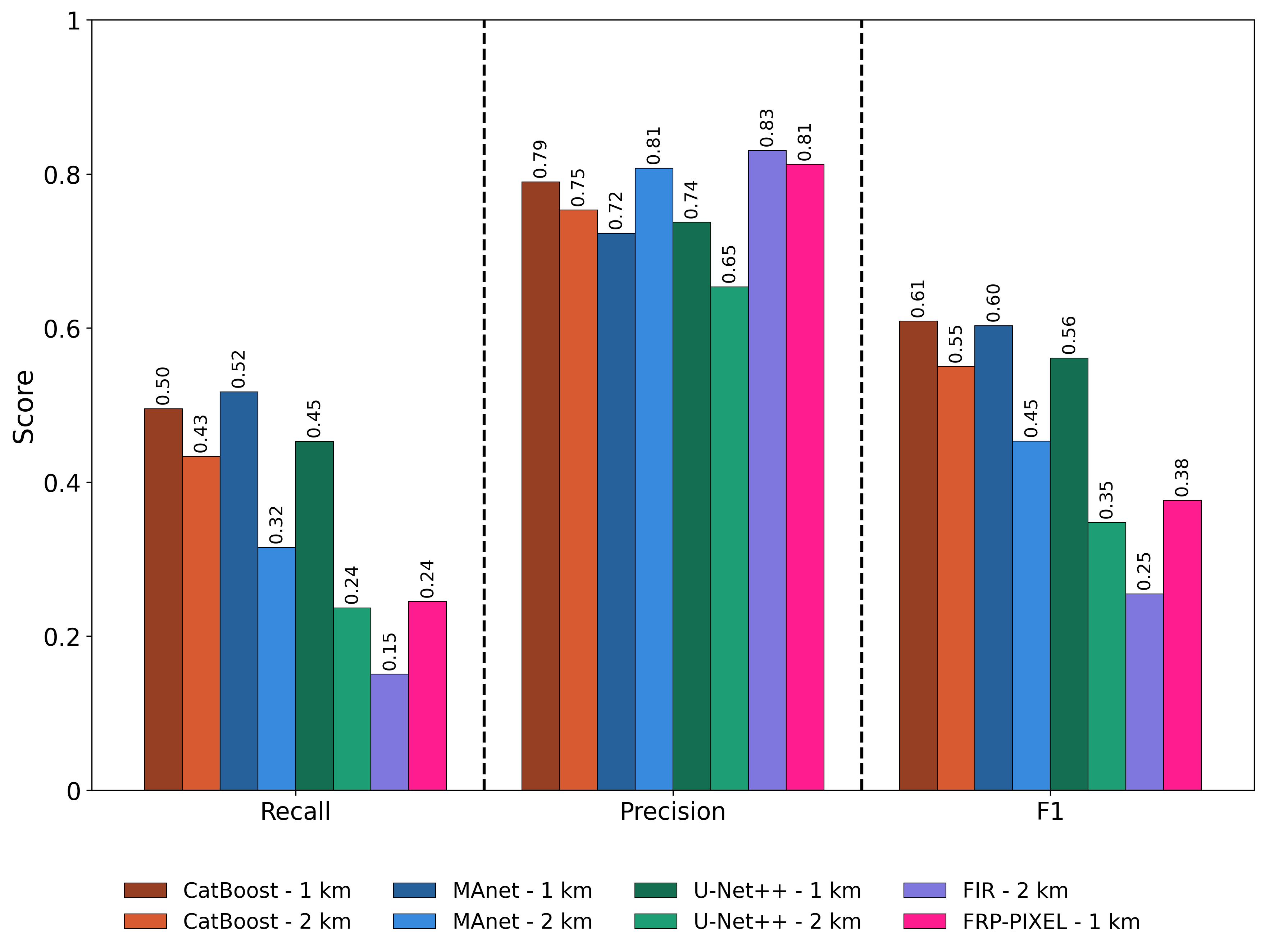}
    \caption{Overall recall, precision, and F1 scores aggregated over all evaluation regions for the 1-km and 2-km models alongside the FIR and FRP-PIXEL.}
    \label{fig:overall_performance}
\end{figure}

As shown in Figure~\ref{fig:overall_performance}, the 1-km CatBoost and DL architectures yielded the highest overall F1 scores, outperforming both their 2-km variants and the threshold products. The 1-km models achieved F1 scores ranging from 0.56 to 0.61, representing a notable gain over FIR (0.25) and FRP-PIXEL (0.38). The 1-km models perform better primarily due to higher recall. They gain +0.06 to +0.21 in recall over 2-km variants, and up to +0.37 over the threshold-based benchmarks.

The threshold-based products achieved high precision scores of 0.83 for FIR and 0.81 for FRP-PIXEL. Conversely, 1-km models have slightly lower values ranging from 0.72 to 0.79. This difference in precision occurs because the threshold-based products are more conservative. Consequently, they fail to detect many true fires. For instance, in Germany, reference VIIRS detections intersected 1,569 MTG FCI 1-km grid cells and 1,367 2-km grid cells. However, FRP-PIXEL detected 65 fire pixels, while FIR identified only 36. In contrast, the 1-km MAnet, U-Net++, and CatBoost models identified 487, 604, and 197 fire pixels, respectively.

\begin{figure}
    \centering
    \includegraphics[width=.7\textwidth]{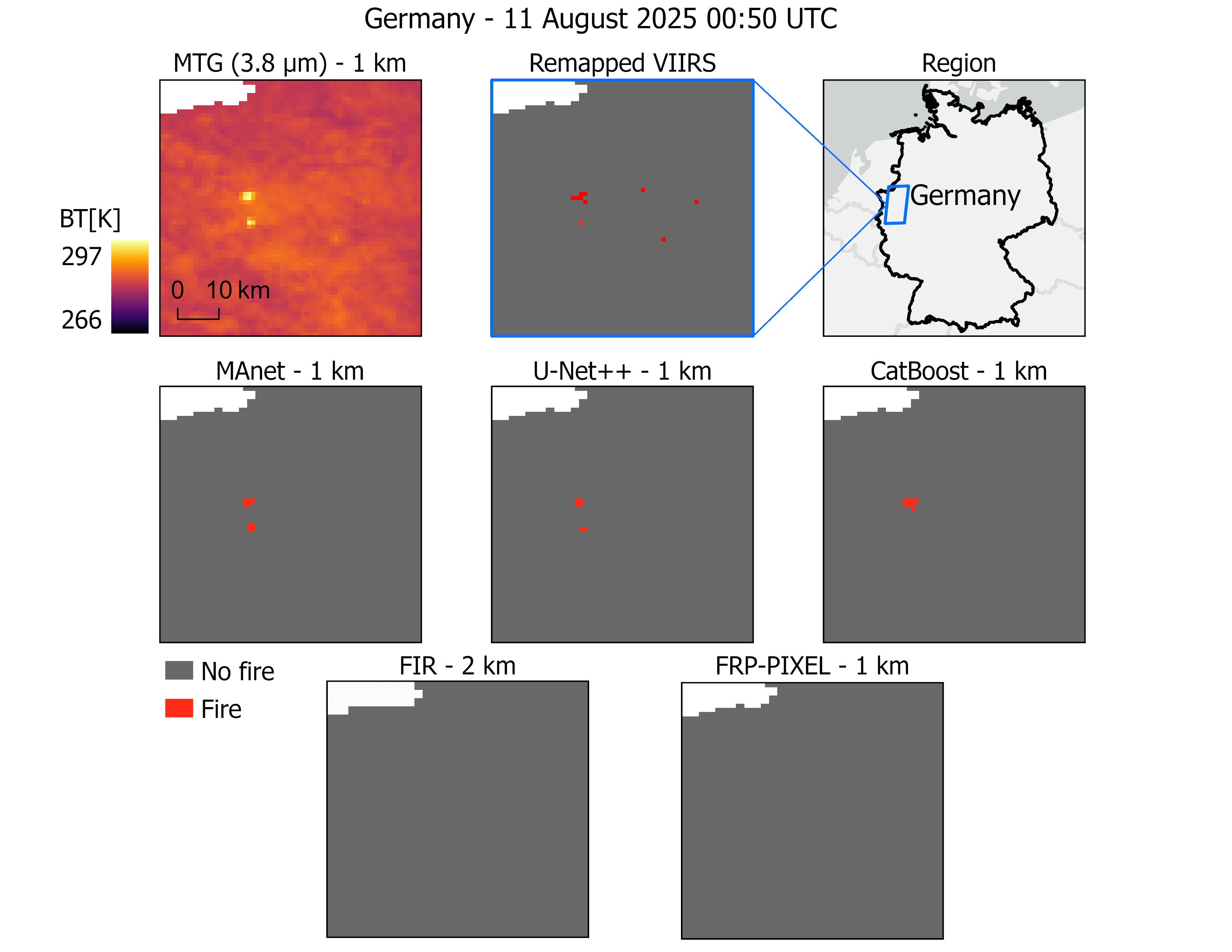}
    \caption{Comparison of fire detection outputs in Germany on 11 August 2025 (00:50 UTC). The panels display predictions from the 1-km MAnet, U-Net++, and CatBoost models, alongside the 1-km FRP-PIXEL and 2-km FIR. Remapped VIIRS fire detections (00:48 UTC) are provided as reference data.}
    \label{fig:germany_example}
\end{figure}

Figure~\ref{fig:germany_example} illustrates the superior detection capabilities of the 1-km ML models compared to the threshold products. It displays an MTG FCI scene acquired on 11 August 2025 at 00:50 UTC alongside corresponding VIIRS reference detections from 00:48 UTC. The 1-km models successfully detect fires visible in the 3.8~$\mu\text{m}$ band that the FIR and FRP-PIXEL miss.

\subsection{Probability of Fire Detection}\label{subsec:probability_of_detection}

This section evaluates model detection probabilities as a function of the number of VIIRS fire pixels within a $3\times3$ window. Let $A_{out}$ denote the output of a fire detection algorithm, where a value of 1 indicates a detected fire and 0 otherwise. For a window parameter $w$ (where $w=1$ corresponds to a $3 \times 3$ window), the neighborhood set $NE_{ij}$ around pixel $(i,j)$ is defined as:
\begin{equation} \label{eq:neighborhood}
NE_{ij} = \{ (i', j') \mid \max(|i' - i|, |j' - j|) \leq w \}
\end{equation}

For various values of an integer $k$, the total number of TP pixels, $NE_{TP,k}$, was calculated. This metric represents pixels with $k$ VIIRS fire detections that also contain a model fire detection within distance $w$. For each output $A_{out}$, let $V_{rm}$ represent the corresponding remapped VIIRS output with threshold $k$:
\begin{equation} \label{eq:truepos}
NE_{TP,k} = \sum_{\text{images}} \sum_{(i,j)} \mathbf{1} \left[ V_{rm}(i,j) = k \text{ and for some } (i', j') \in NE_{ij}, \; A_{out}(i', j') = 1 \right]
\end{equation}
Similarly, the number of FN pixels, $NE_{FN,k}$, was calculated:
\begin{equation} \label{eq:falseneg}
NE_{FN,k} = \sum_{\text{images}} \sum_{(i,j)} \mathbf{1} \left[ V_{rm}(i,j) = k \text{ and for all } (i', j') \in NE_{ij}, \; A_{out}(i', j') = 0 \right]
\end{equation}
The sum $NE_{TP,k} + NE_{FN,k}$ represents the total number of pixels containing $k$ VIIRS fire detections. The probability of a model to detect a fire within a window $w$ around each such pixel is estimated by:
\begin{equation} \label{eq:probability}
P \left( A_{out}(i', j') = 1 \text{ for at least one } (i', j') \in NE_{ij} \mid V_{rm}(i,j) = k \right) = \frac{NE_{TP,k}}{NE_{TP,k} + NE_{FN,k}}
\end{equation}

\begin{figure}
    \centering
    \includegraphics[width=.5\textwidth]{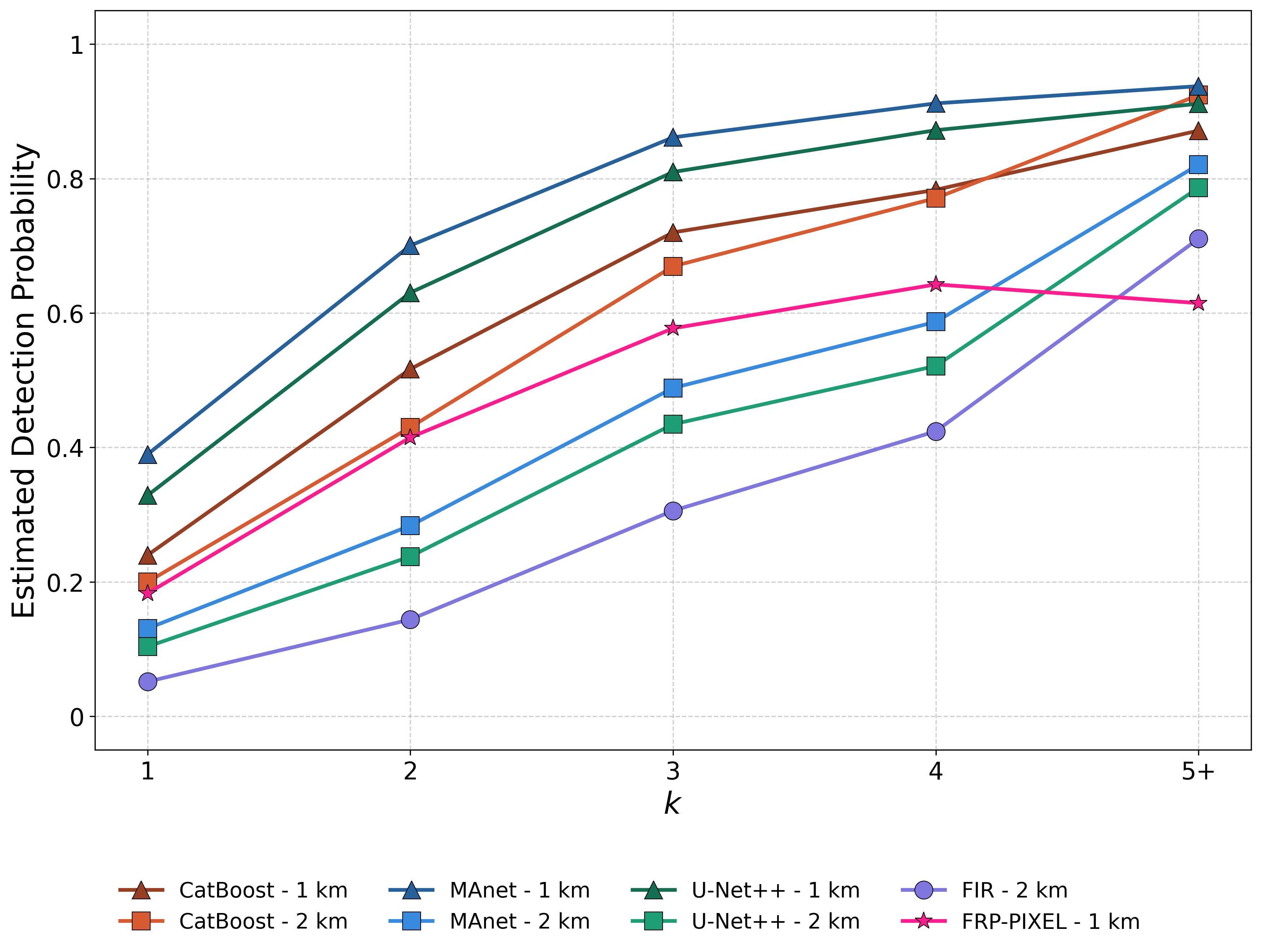}
    \caption{Probability of fire pixel detection by ML, FIR, and FRP-PIXEL models as a function of the number of VIIRS fire detections ($k$) inside an MTG FCI pixel.}
    \label{fig:fire_detection_probability}
\end{figure}

Figure~\ref{fig:fire_detection_probability} presents the estimated detection probabilities aggregated over all evaluation regions as a function of $k$. Here, $k$ represents the number of VIIRS fire detections contained within a single MTG FCI pixel. A region-specific probability analysis is provided in Appendix~\ref{app:regional_probability}.

For all models, detection probability increases with $k$, confirming that more VIIRS detections within an MTG FCI pixel yield higher detection rates. The 1-km DL architectures consistently achieve the highest detection probabilities for all values $k$. They reach a probability of 0.33–0.39 for small fires ($k=1$) and exceed 0.91 for large fires ($k \ge 5$). The 1-km and 2-km CatBoost models follow closely, with a slightly lower detection probability.

In contrast, threshold-based products exhibited considerably lower probabilities. For small fires ($k=1$), FRP-PIXEL and FIR reach detection probabilities of only 0.18 and 0.05, respectively. Notably, while FIR sensitivity increase steadily with $k$, reaching 0.71 at $k \ge 5$, the FRP-PIXEL curve plateau around $k \ge 3$. This plateau may occur because fire pixels with $k \ge 3$ belong to larger fire clusters. While the cluster itself is detected by the FRP-PIXEL, not all VIIRS pixels are within the $3 \times 3$ window. Expanding the spatial window to $5 \times 5$ could yield higher detection probabilities in such cases. Overall, the 1-km ML models demonstrated greater sensitivity to both small fires ($k=1$) and large fire events ($k \ge 5$). They achieved higher detection probabilities than both their 2-km variants and the threshold products.

\subsection{Early Detection Analysis}\label{subsec:early_detection}

Timely fire detection is critical for preventing small fires from escalating into large, uncontrollable events. Having established the superior overall detection capabilities of the 1-km CatBoost, MAnet, and U-Net++ models, next, their early fire detection performance is compared to the FIR and FRP-PIXEL products. This evaluation used a database of 20 distinct fire events (Section \ref{subsubsec:fire_events_databse}).To focus the evaluation on the highest-performing configurations, only the 1-km variants were analyzed. Each event was evaluated over a 10-hour temporal window spanning from 8 hours before to 2 hours after the initial VIIRS fire detection.

\begin{figure}
    \centering
    \includegraphics[width=.9\textwidth]{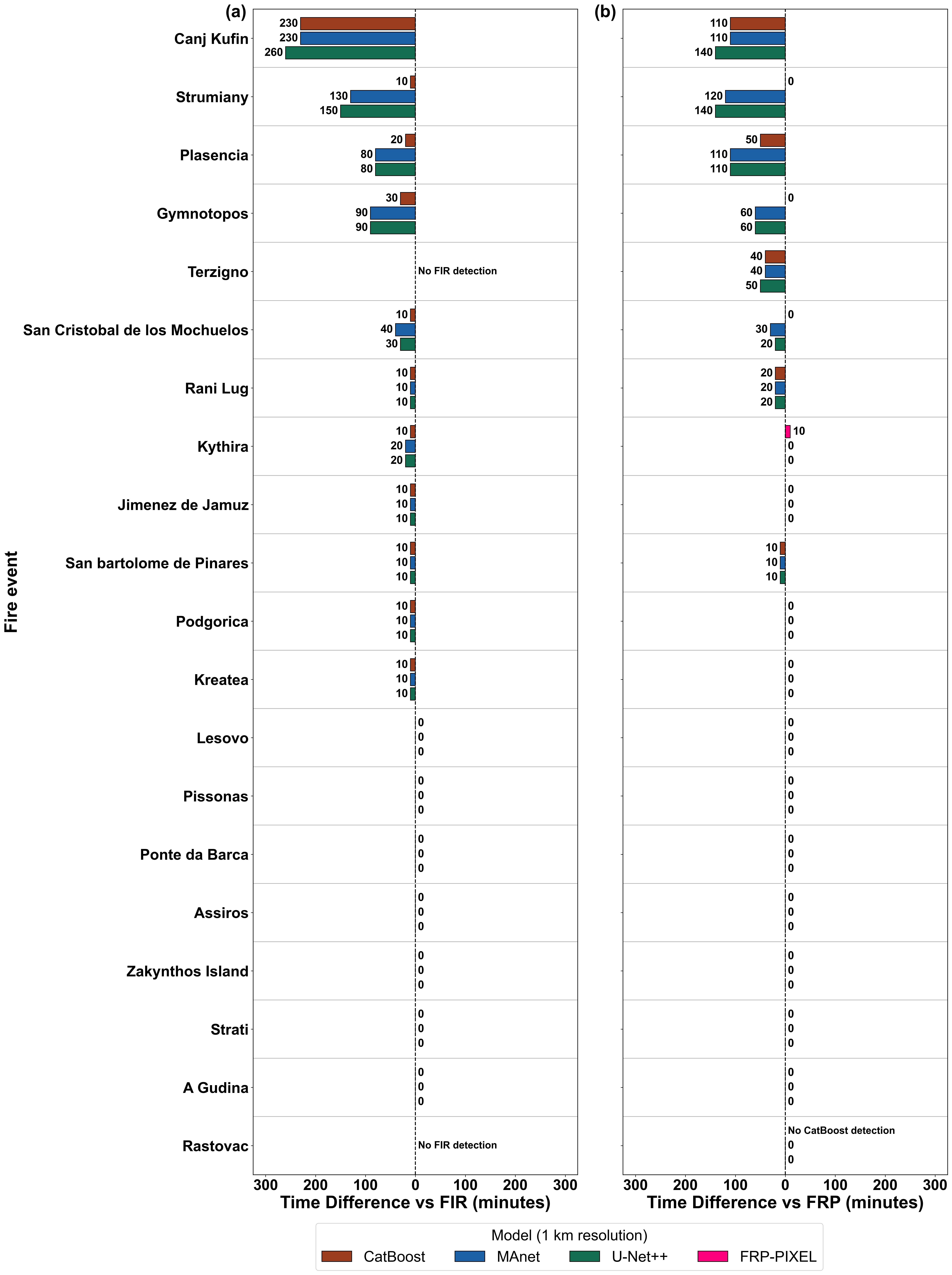}
    \caption{Time difference in initial fire detection between the 1-km ML models and (a) FIR, and (b) FRP-PIXEL}
    \label{fig:early_detection_all_fires}
\end{figure}

Figure~\ref{fig:early_detection_all_fires} summarizes the early detection lead times ($\Delta t$, in minutes) of the 1-km models compared to FIR (Figure~\ref{fig:early_detection_all_fires}a) and FRP-PIXEL (Figure~\ref{fig:early_detection_all_fires}b). A lead time of zero indicates simultaneous detection. Fire events missed by a given model are marked accordingly. 

To quantify and compare early detection, several key metrics were computed based on the counts of:
early ($N_{\text{win}}$), simultaneous ($N_{\text{same}}$), and missed/late ($N_{\text{miss}}$) detections across the all events ($N_{\text{total}}$). The probability of detecting a fire earlier than the threshold products, $P_{\text{win}}$, is defined as:
\begin{equation}
    P_{\text{win}} = \frac{N_{\text{win}}}{N_{\text{total}}} \label{eq:p_win}
\end{equation}
Similarly, the probability of simultaneous detection, $P_{\text{same}}$, is given by:
\begin{equation}
    P_{\text{same}} = \frac{N_{\text{same}}}{N_{\text{total}}} \label{eq:p_same}
\end{equation}
To evaluate lead time for successful early detections ($\Delta t > 0$), the conditional expected lead time, $\mathbb{E}[\Delta t \mid \text{win}]$, was calculated in minutes as:
\begin{equation}
    \mathbb{E}[\Delta t \mid \text{win}] = \frac{1}{N_{\text{win}}} \sum_{i=1}^{N_{\text{win}}} \Delta t_i \label{eq:cond_exp}
\end{equation}
where $\Delta t_i$ represents the lead time in minutes for event $i$. Table~\ref{tab:early_detection_stats}summarizes these metrics for all evaluated models.

\begin{table}
    \centering
    \caption{Early detection metrics comparing 1-km models to the FIR and FRP-PIXEL. The metrics include total early detection counts ($N_{\text{win}}$), simultaneous detections ($N_{\text{same}}$), missed detections ($N_{\text{miss}}$), corresponding probabilities ($P_{\text{win}}$ and $P_{\text{same}}$), and the conditional expected lead time ($\mathbb{E}[\Delta t \mid \text{win}]$) in minutes.}
    \label{tab:early_detection_stats}
    \begin{tabular}{lcccccc}
        \hline
        \textbf{Total Fire Events} & \multicolumn{6}{c}{\textbf{20}} \\
        \hline
        & \multicolumn{3}{c}{\textbf{vs. FIR (2-km)}} & \multicolumn{3}{c}{\textbf{vs. FRP-PIXEL (1-km)}} \\
        \cmidrule(lr){2-4} \cmidrule(lr){5-7}
        \textbf{Metric} & \textbf{CatBoost} & \textbf{MAnet} & \textbf{U-Net++} & \textbf{CatBoost} & \textbf{MAnet} & \textbf{U-Net++} \\
        \hline
        Earlier Detections ($N_{\text{win}}$) & 12 & 13 & 13 & 4 & 8 & 8 \\
        Same Time Detections ($N_{\text{same}}$) & 7 & 7 & 7 & 14 & 12 & 12 \\
        Missed / late Events ($N_{\text{miss}}$) & 0 & 0 & 0 & 2 & 0 & 0 \\
        Probability of Win ($P_{\text{win}}$) & 0.63 & 0.65 & 0.65 & 0.20 & 0.40 & 0.40 \\
        Probability of Same Time ($P_{\text{same}}$) & 0.37 & 0.35 & 0.35 & 0.70 & 0.60 & 0.60 \\
        \textbf{Conditional Expected Value $\mathbb{E}[\Delta t \mid \text{win}]$ (minutes)} & \textbf{32.73} & \textbf{58.18} & \textbf{61.82} & \textbf{45} & \textbf{65} & \textbf{70} \\
        \hline
    \end{tabular}
\end{table}

Compared to the FIR product, all three 1-km models detected fires earlier in most events ($P_{\text{win}} = \sim0.65$), with simultaneous detections in the remaining cases ($P_{\text{same}} = \sim0.35$). When early detection occurred, the DL architectures yielded substantial conditional lead times, averaging 58.18 minutes for MAnet and 61.82 minutes for U-Net++, compared to 32.73 minutes for CatBoost. 

Compared to FRP-PIXEL, early detection probabilities decreased for all ML models, particularly for CatBoost ($P_{\text{win}} = 0.20$). This reduction in early detection aligns with the overall performance findings (Section~\ref{subsec:overall_performance}), where FRP-PIXEL achieved higher recall and F1 scores compared to FIR. Nevertheless, when the DL models detected fires prior to FRP-PIXEL, the conditional lead times remained substantial, averaging 65 minutes for MAnet and 70 minutes for U-Net++. 

\begin{figure}
    \centering
    \includegraphics[width=.95\textwidth]{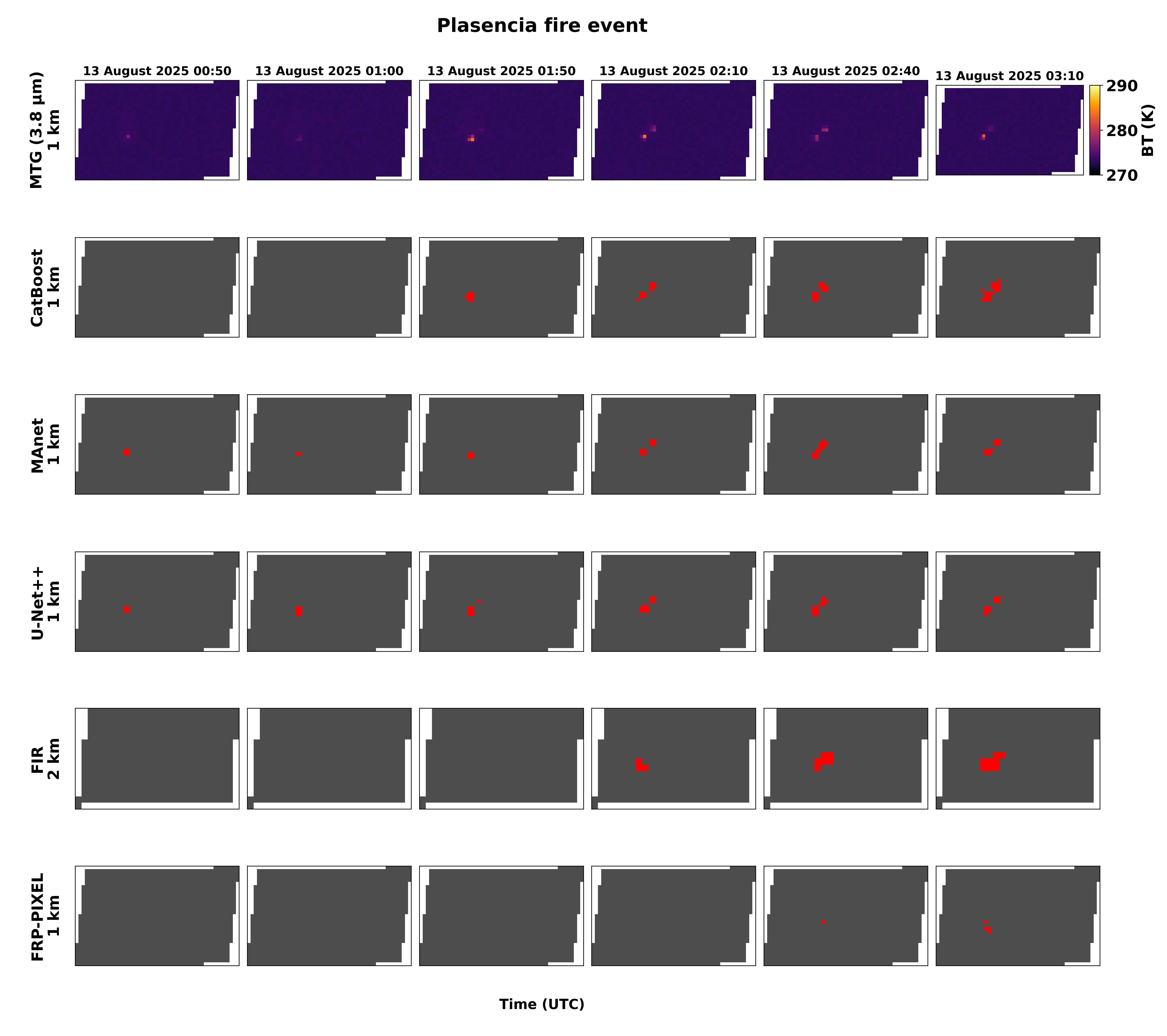}
    \caption{Sequence of detections for the Plasencia fire event. Illustrating early detection by the MAnet and U-Net++ models prior to CatBoost, FIR, and FRP-PIXEL.}
    \label{fig:Plasencia_fire}
\end{figure}

An example of the ML models' early-detection capability is illustrated in the Plasencia fire event in Spain (Figure~\ref{fig:Plasencia_fire}). MAnet and U-Net++ detected the fire on 13 August 2025 at 00:50 UTC, identifying it 60 minutes prior to CatBoost, 80 minutes prior to FIR, and nearly two hours prior to FRP-PIXEL. Overall, neither FIR nor FRP-PIXEL detected a fire before the DL models, whereas FRP-PIXEL preceded CatBoost in only two events.

\clearpage

\section{Discussion and Conclusions}\label{sec:discussion}

This study compares the performance of ML and threshold-based models for fire detection using MTG FCI imagery. Two distinct datasets of 1-km HR and 2-km NR imagery were constructed from over 1,700 MTG FCI images. These covered Eastern Africa, Western Europe, and the Eastern Mediterranean. Reference annotations were derived from VIIRS fire product, taking into account spatial misalignments (Section~\ref{subsec:dataset_construction}). To evaluate generalization, models were tested on an independent dataset of over 600 MTG FCI scenes. This test set covered regions outside the training area, including Eastern Europe, the Middle East, and Western Africa.

The overall evaluation indicates that the 1-km MAnet and U-Net++ models outperform their 2-km variants and both benchmark products. They consistently achieve higher recall and F1 scores in all evaluated regions. Furthermore, the 1-km DL models demonstrate higher detection probabilities for small and low-intensity fires (Section~\ref{subsec:probability_of_detection}).

High recall and F1 scores were maintained in unseen test regions, demonstrating strong model robustness and generalization. For instance, in the Balkan region (Appendix \ref{app:regional_performance}), 1-km U-Net++ achieved an F1 score of 0.52 compared to $\sim0.35$ for FIR and FRP-PIXEL. Furthermore, in Nigeria, 1-km MAnet achieved an F1 score of 0.57 versus 0.12 for FIR and 0.34 for FRP-PIXEL. Region-specific analysis revealed that model accuracy depends on fire characteristics. In regions with large, high-intensity wildfires such as Spain and Portugal, all models achieved F1 scores exceeding 0.65. Conversely, in regions with low-intensity fires, such as Germany, F1 scores dropped below 0.41 for all models.
In terms of precision, the 1-km ML models achieved scores slightly lower than those of the threshold-based products. However, as noted in Section~\ref{subsec:overall_performance}, the threshold-based products are more conservative than the CatBoost and DL models, leading to missed detections.

Spatial resolution differences were also prominent. The 2-km DL models exhibited lower F1 scores than their 1-km counterparts. This drop is primarily driven by lower recall, as coarser-resolution models fail to detect small fires. In contrast, the 2-km CatBoost model achieved performance comparable to its 1-km variant. This result demonstrates the robustness of gradient boosting when trained on limited datasets. Among the evaluated DL architectures, the hybrid CNN-Transformer MAnet outperformed the CNN-based U-Net++. This highlights the advantage of pairing CNNs for local pattern extraction with Transformers for global context.

Our FRP-PIXEL accuracy metrics differ slightly from the results reported by \citet{XU_2026_MTG_FCI_fire_improve}. They reported an F1 score of 0.47, whereas we obtained 0.38. This divergence can stem from several methodological differences. \citet{XU_2026_MTG_FCI_fire_improve} evaluated a 16-day period in September 2024, restricted to Southern Africa. In contrast, our evaluation spans multiple months across 2025 and 2026 in different geographical regions. Additionally, \citet{XU_2026_MTG_FCI_fire_improve} applied a $7 \times 7$ pixel buffer window to account for spatial misalignment, whereas this study used a $3 \times 3$ window. Notably, the F1 scores of our 1-km ML models are higher by 0.09-0.14 than those reported by \citet{XU_2026_MTG_FCI_fire_improve}.

The results presented in this study are in agreement with previous work on ML fire detection using GEO imagery. For instance, \citet{KANG_2023_Toward_H8} trained a CNN model on Himawari-8 AHI imagery and reported F1 scores of 0.50–0.66 in Asia. \citet{Zhao_2022_DL_GOES} trained and evaluated a Recurrent Neural Network (RNN) model on GOES ABI imagery for various large-scale fire events. They reported F1 scores ranging from 0.18 to 0.71. 
Furthermore, \citet{Mercatini_MTG_FCI_fire_detection_2026} evaluated a U-Net model on MTG FCI imagery over Italy. They reported an F1 score of 0.76. 

Beyond overall detection metrics, the early detection analysis (Section \ref{subsec:early_detection}) demonstrates clear lead-time advantages for the 1-km ML models. Neither FIR nor FRP-PIXEL detected fires before the 1-km DL models. Furthermore, FRP-PIXEL preceded CatBoost in only two events. The DL and CatBoost models frequently detect fires simultaneously with FIR and FRP-PIXEL. However, when ML models detect fires first, they provide a substantial lead time ranging from 32.73 to 70 minutes. Looking ahead, MTG FCI imagery will become available at higher temporal resolutions of 2.5 and 5 minutes. This frequent acquisition could potentially improve the early detection of the 1-km ML models.

While these findings highlight the potential of ML for MTG FCI fire detection, several limitations should be addressed in future work.
First, the dataset lacks coverage at high latitudes where the pixel is larger. Evaluating high-latitude areas can help determine the effects of large viewing angles on model accuracy. Second, our precision score was slightly lower than that of the threshold-based algorithms. This leaves room to reduce false alarms in future model fine-tuning. Third, training and evaluation datasets are from the same calendar year. Third, our current dataset relies on a single calendar year. Training on a multi-year dataset would capture broader climatic variations and could improve model accuracy and generalization. Finally, reference labels were derived exclusively from VIIRS. Integrating higher-resolution satellite imagery (e.g., Landsat or Sentinel-2) or fire agency reports could improve future model training and evaluation.

In summary, this study presents a practical application of ML models for MTG FCI fire detection. The findings demonstrate that 1-km HR models significantly outperform both 2-km NR models and threshold-based algorithms in terms of overall accuracy. Importantly, these models can detect early-stage fires up to an hour faster than threshold products. To support future research and application, all trained models are publicly accessible via a GitHub repository \footnote{\url{https://github.com/asafvanunu/Github_repo_MTG_ML_DL_fire_detection}}.

\section*{Funding}
This work was supported by the Israel Space Agency, which is part of the Ministry of Science and Technology, Israel, under Grant 3-18142; The Jewish National Fund (JNF) under Grant 2006/22, and the Institute for Environmental Sustainability (IES) at the Weizmann Institute of Science

\printcredits

\bibliographystyle{cas-model2-names}

\bibliography{refs}

\clearpage
\appendix
\counterwithin{figure}{section}
\renewcommand{\thefigure}{\thesection.\arabic{figure}}

\counterwithin{table}{section}
\renewcommand{\thetable}{\thesection.\arabic{table}}

\section{Fire Events Summary}
\label{app:fire_events_list}

Table \ref{tab:fire_events} details each fire event ID, Copernicus EMS ID, EMS Area of Interest (AOI) number, and event name. Additionally, it reports the time of the initial VIIRS detection, the country of occurrence, the centroid coordinates (longitude and latitude), and the total fire size (in km\textsuperscript{2}).

\begin{table}
\centering 
\caption{Summary of 20 fire events used for first-time detection comparison. The table lists the study ID, EMS ID, EMS AOI, fire event name, timestamp of first VIIRS detection, country, centroid coordinates (longitude and latitude) and fire size (in km\textsuperscript{2}).}
\vspace{0.5em}
\label{tab:fire_events} 
\resizebox{\textwidth}{!}{%
\begin{filecontents*}{figs/raw_csv.csv}
ID,EMS ID,EMS AOI,Fire event,VIIRS first detection time (UTC),Country,Longitude,Latitude,Fire size (km2)
1,EMSR836,6,Canj Kufin,10/8/2025 0:48,Montenegro,19.001,42.187,53
2,EMSR822,2,Strumiany,26/7/2025 0:30,Bulgaria,23.26,41.684,195
3,EMSR837,7,Plasencia,13/8/2025 3:12,Spain,-5.902,40.178,562
4,EMSR835,1,Gymnotopos,11/8/2025 12:12,Greece,20.889,39.254,800
5,EMSR830,1,Terzigno,6/8/2025 0:24,Italy,14.467,40.815,64
6,EMSR837,17,San Cristobal de los Mochuelos,15/8/2025 12:42,Spain,-6.222,41.076,513
7,EMSR822,1,Rani lug,27/7/2025 0:12,Bulgaria,22.539,42.814,221
8,EMSR818,1,Kythira,26/7/2025 10:36,Greece,22.97,36.227,204
9,EMSR837,5,Jimenez de Jamuz,10/8/2025 12:36,Spain,-6.026,42.179,1818
10,EMSR837,10,San Bartolome di Pinares,8/8/2025 12:48,Spain,-4.339,40.56,76
11,EMSR836,1,Podgorica,11/8/2025 10:36,Montenegro,19.309,42.501,451
12,EMSR828,1,Keratea,8/8/2025 11:30,Greece,23.994,37.723,132
13,EMSR822,3,Lesovo,27/7/2025 0:12,Bulgaria,26.393,42.024,994
14,EMSR819,1,Pissonas,26/7/2025 11:54,Greece,23.703,38.481,313
15,EMSR824,2,Ponte da Barca,27/7/2025 1:54,Portugal,-8.258,41.817,368
16,EMSR825,1,Assiros,29/7/2025 23:36,Greece,23.066,40.826,457
17,EMSR832,1,Zakynthos Island,12/8/2025 11:30,Greece,20.791,37.723,221
18,EMSR823,1,Strati,27/7/2025 0:12,Greece,23.739,35.282,259
19,EMSR837,12,A Gudina,11/8/2025 2:12,Spain,-7.27,42.176,1857
20,EMSR836,5,Rastovac,11/8/2025 0:30,Montenegro,18.928,42.867,108
\end{filecontents*}
\csvreader[
    tabular=|>{\centering\arraybackslash}p{0.5cm}|c|c|c|c|c|c|c|c|,
    table head=\hline \centering ID & 
    \begin{tabular}{@{}c@{}}EMS\\ ID \end{tabular}& 
    \begin{tabular}{@{}c@{}}EMS\\ AOI \end{tabular}& 
    \begin{tabular}{@{}c@{}}Fire\\ event \end{tabular}& 
    \begin{tabular}{@{}c@{}}VIIRS first\\ detection time\\(UTC)\end{tabular} & 
    Country & 
    Longitude & 
    Latitude & 
    \begin{tabular}{@{}c@{}}Fire size\\(km\textsuperscript{2})\end{tabular} 
    \\\hline,
    table foot=\hline
]{figs/raw_csv.csv}{}%
{\raggedright\csvcoli & \csvcolii & \csvcoliii & \csvcoliv & \csvcolv & \csvcolvi & \csvcolvii & \csvcolviii & \csvcolix}
}
\end{table}

\section{CatBoost Hyperparameters}
\label{app:CatBoost Hyperparameters}
\begin{figure}
    \centering
    \includegraphics[width=.9\textwidth]{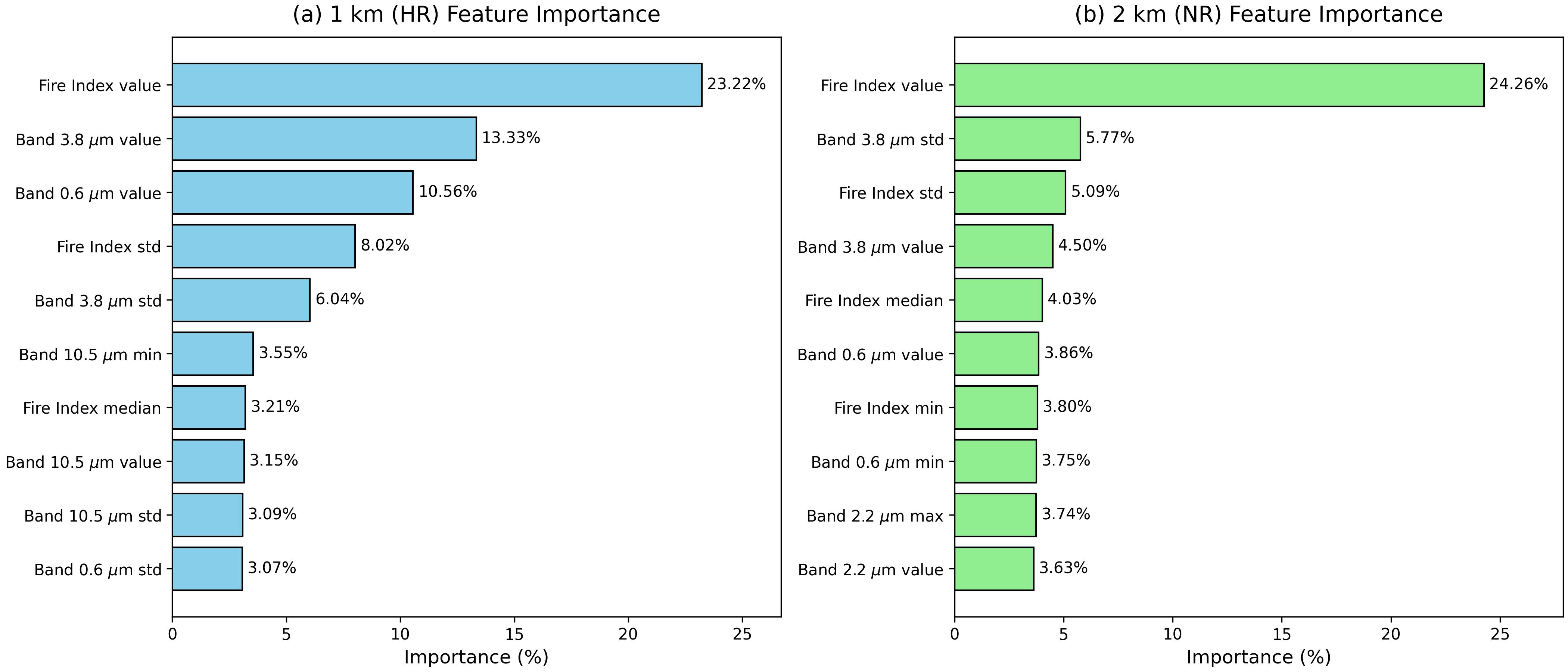}
    \caption{Ten most important features for the (a) 1-km (HR) model and (b) 2-km (NR) model.}
    \label{fig:feature importance}
\end{figure}

The CatBoost algorithm has several key hyperparameters to tune its tree-building process.
The \texttt{learning rate} dictates the step size taken during gradient descent optimization; smaller values, such as 0.01, ensure training stability. 
The number of \texttt{iterations} (set to 1000) determines the total number of decision trees constructed. A higher count enables the model to learn more intricate patterns from the data. The structural complexity of individual trees is managed by the \texttt{depth} parameter. It controls the maximum number of splits per tree. This parameter is configured to 6 for the 1-km model and 8 for the 2-km model.
To mitigate overfitting, two regularization techniques were applied. Setting \texttt{L2 leaf regularization} was set to 5 to penalize large leaf output values and prevent disproportionate prediction updates. Additionally, the \texttt{bagging temperature} was set to 5 to introduce stochasticity to weights to prevent overfitting. These hyperparameters were selected using a 5-fold cross-validation grid search with the Logloss objective function. The search optimized specifically for precision to minimize false alarms.
Figure~\ref{fig:feature importance}(a) and Figure~\ref{fig:feature importance}(b) display the ten most important features for the 1-km HR and 2-km NR models, respectively. Several of the highest-ranked features are derived from the FI and its statistics. This includes the standard deviation of FI from neighboring pixels that capture local variability. Additionally, the 3.8$\mu$m MIR band, which is sensitive to fires, has high relative importance. Notably, the 0.6$\mu$m VIS band also has high importance within the 1-km HR model. Suggesting that the algorithm uses reflectance to distinguish between daytime and nighttime illumination conditions.

\section{DL Models Hyperparameters}
\label{app:DL Hyperparameters}
This section provides the network configurations and hyperparameters used to train the MAnet with MiT backbone and U-Net++ with SE-ResNet backbone. To ensure there was no potential data leakage, the dataset was randomly split into 80\% training and 20\% validation before computing global descriptive statistics. Input image patches were normalized channel-wise using min-max scaling. The scaling parameters were derived exclusively from the training dataset. To handle class imbalance, a \texttt{WeightedRandomSampler} was used to balance patch selection. In addition, data augmentations were applied to fire-containing patches during training. These included horizontal flips ($p = 0.5$), vertical flips ($p = 0.5$), and random 90-degree orthogonal rotations ($p = 0.5$). Both models were initialized using ImageNet pre-trained encoder weights and optimized end-to-end. Models were trained with a batch size of 32 using the AdamW optimizer. Both the initial learning rate ($\eta$) and weight decay coefficient were set to $10^{-4}$. The learning rate was dynamically adjusted across the 50 training epochs using a Cosine Annealing learning rate scheduler ($T_{\max} = 50$). To optimize performance under class imbalance, a Focal Loss objective function was integrated with a positive class upweighting coefficient. The focal focusing parameter ($\gamma$) was set to 2.0 to down-weight background pixels, and the positive class multiplier weight ($w$) was set to 10 to penalize missed fire pixels. Regularization was reinforced by clipping gradient norms at a maximum threshold of 1.0. Model checkpoints were evaluated at the end of each epoch using the validation subset. Performance was tracked via the fire-class Intersection over Union (IoU) metric at a baseline sigmoid threshold of $\beta = 0.5$. An stopping mechanism monitored validation IoU with a patience of 15 epochs. If performance failed to improve, optimization halted and the highest-performing weights were reloaded.

\section{Regional Model Performance}
\label{app:regional_performance}

\begin{figure}
    \centering
    \includegraphics[width=.9\textwidth]{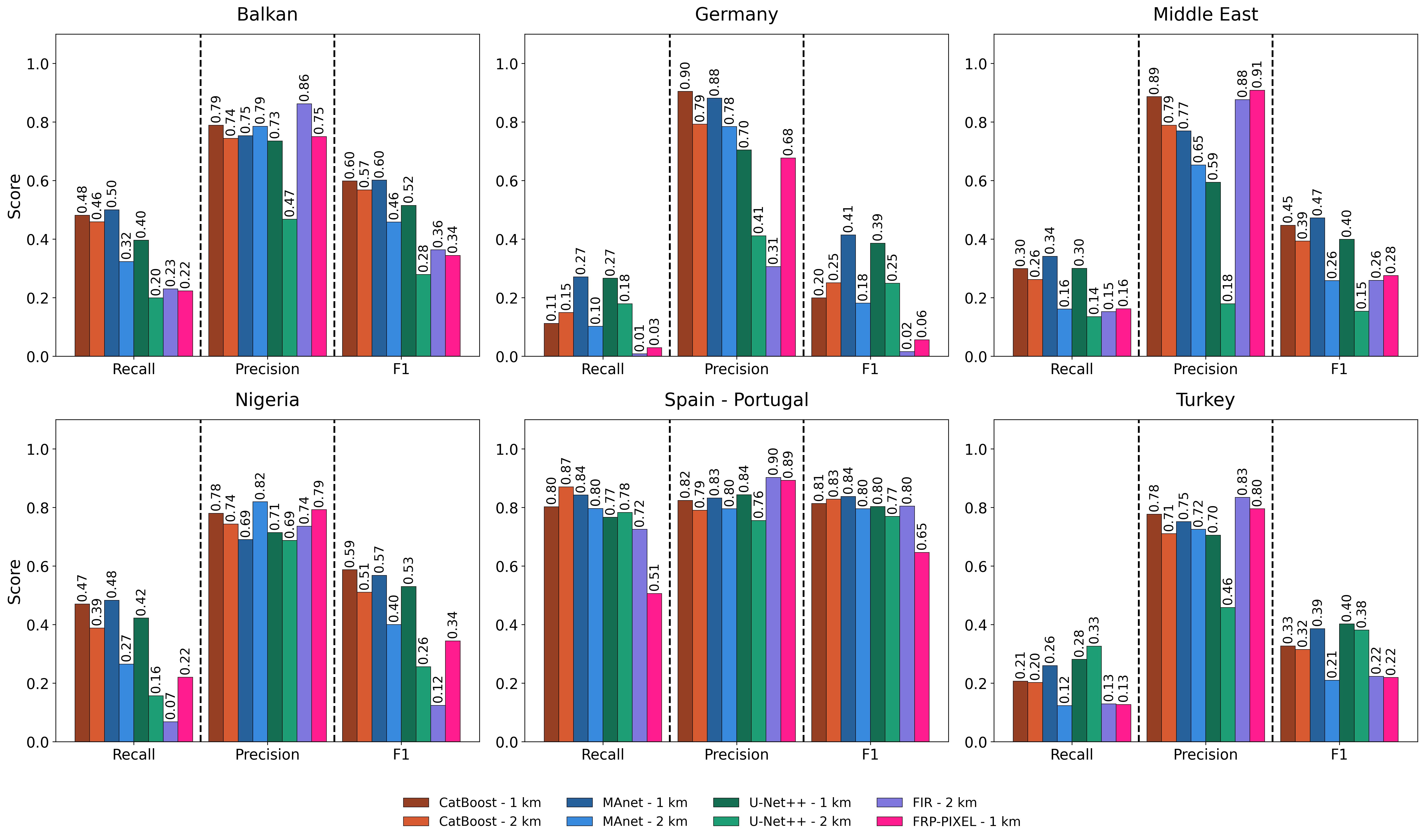}
    \caption{Comparison of recall, precision, and F1 scores across all evaluation regions for the 1 and 2-km ML models alongside the FIR (2-km) and FRP-PIXEL (1-km) benchmark products.}
    \label{fig:regional_performance}
\end{figure}

This appendix presents the regional performance breakdown across all six evaluated study areas, complementing the aggregated global metrics reported in Section~\ref{subsec:overall_performance}. Figure~\ref{fig:regional_performance} illustrates the recall, precision, and F1 scores across individual regions.

Across nearly all study areas, the 1-km MAnet and U-Net++ models consistently outperformed their 2-km variants and the benchmark products. For example, in the Middle East, 1-km MAnet achieved an F1 score of 0.47, surpassing FRP-PIXEL, FIR and the 2-km variant by roughly 0.20. Similarly, 1-km U-Net++ maintained an F1 lead over benchmark products in almost all regions. The strong performance of the 1-km ML models was driven primarily by higher recall values.

In contrast to the DL models, CatBoost exhibited stability in 1 and 2-km configurations. Both CatBoost models consistently outperformed FIR and FRP-PIXEL across all regions, with slight differences in performance. Furthermore, 2-km CatBoost frequently outperformed the 2-km MAnet and U-Net++.

The high FIR and FRP-PIXEL precision scores corresponded to very low detection counts. In Turkey, VIIRS fire detections intersected 3,714 1-km and 3,089 2-km MTG FCI grid cells. FRP-PIXEL and FIR identified only 584 and 493 fire pixels, respectively. In contrast, the 1-km MAnet, U-Net++, and CatBoost models identified 1,255, 1,514, and 1,063 fire pixels, respectively.

\section{Regional Fire Detection Probability}
\label{app:regional_probability}

\begin{figure}
    \centering
    \includegraphics[width=.9\textwidth]{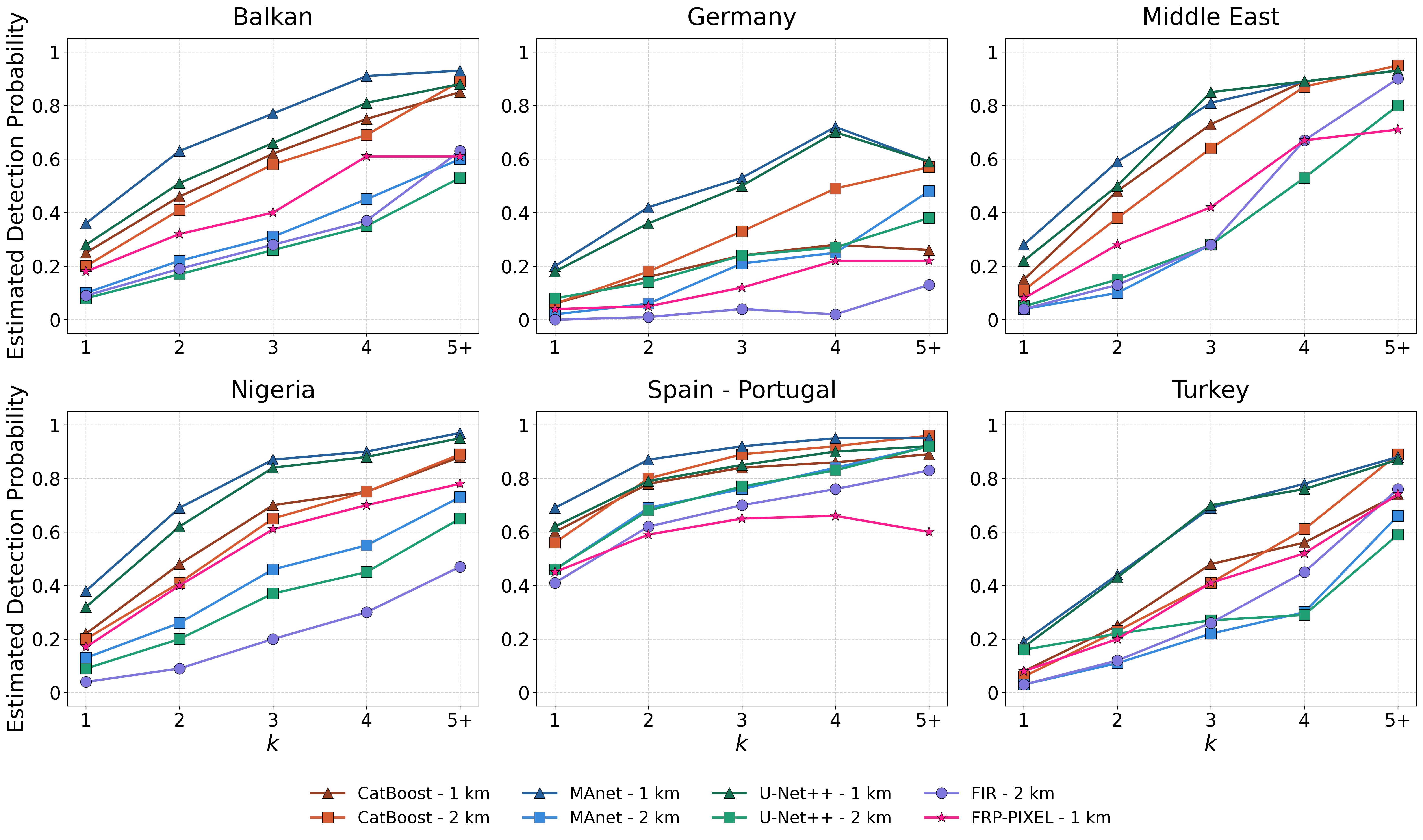}
    \caption{Region-specific probability of fire detection as a function of the number of VIIRS fire detections per MTG FCI pixel ($k$).}
    \label{fig:regional_probability}
\end{figure}

This appendix provides the regional analysis of fire detection probabilities for the six study areas, complementing the global analysis presented in Section~\ref{subsec:probability_of_detection}. Figure~\ref{fig:regional_probability} details detection probability as a function of VIIRS fire count per MTG FCI pixel ($k$).

In all regions, the 1-km MAnet, U-Net++, and CatBoost models maintained superior detection probabilities for all $k$ values. They consistently outperformed their 2-km variants as well as the FIR and FRP-PIXEL products. The probability gains were most pronounced for small fires ($k=1$).

Detection probability generally increased with $k$ in every individual region. A notable exception occurred in Spain-Portugal for FRP-PIXEL, where detection probability dropped from 0.66 at $k = 4$ to 0.60 at $k \ge 5$. This decline likely stems from large fire clusters extending beyond the $3 \times 3$ evaluation window. Additionally, 1-km DL models in Germany showed a sharp drop at $k \ge 5$, likely due to the rarity of large fires in that region.

\end{document}